\documentclass[preprint,12pt]{elsarticle}
\usepackage{wrapfig}
\usepackage{amssymb}
\usepackage {amsmath}
\usepackage{graphicx}
\usepackage{subfigure}
\usepackage{bm}
\usepackage{booktabs}
\usepackage{threeparttable}
\usepackage{adjustbox}
\usepackage[figuresright]{rotating}
\usepackage{multicol}
\usepackage{multirow}

\usepackage{color}
\newcommand{\tabincell}[2]{\begin{tabular}{@{}#1@{}}#2\end{tabular}}
\usepackage{verbatim}
\usepackage{amsmath, bm}

\journal{Nuclear Physics B}

\begin{document}

\begin{frontmatter}



\title{Computational efficient deep neural network with difference attention maps for facial action unit detection}

 \author[label1]{Jing Chen}\ead{jing\_chen@hrbeu.edu.cn}
 \author[label2]{Chenhui Wang}\ead{harbin0451@g.ucla.edu}
 \author[label1]{Kejun Wang \corref{cor1}}\ead{heukejun@126.com}
 \author[label1]{Meichen Liu}\ead{meichen@hrbeu.edu.cn}
 \address[label1]{College of Intelligent Systems Science and Engineering, Harbin Engineering University, Harbin 150001, China}
\address[label2]{Department of Statistics, UCLA , Los Angeles 90095-1554, California}
\cortext[cor1]{Corresponding author}


\begin{abstract}
There are more and more applications of attention mechanism in the field of computer vision. As a task in the field of vision, facial action unit detection has gradually used attention mechanism. Attention mechanism can make the network pay more attention to specific areas to capture the features related to AU. Most of the existing attention-based methods use facial landmark detection to predefine the regions associated with AU. These methods depend on the accuracy of facial landmark detection. Region learning is not conducive to learning the correlation between AUs. The number of predefined regions of interest is generally fixed. Processing parts that have nothing to do with the target task also increases the computational burden and network complexity. In this paper, we propose a computational efficient end-to-end training deep neural network (CEDNN) model and spatial attention maps based on difference images. Firstly, the difference image is generated by image processing. Then five binary images of difference images are obtained by using different thresholds. These binary images are regarded as spatial attention maps. We use the group convolution to reduce model complexity. Skip connection and $\text{1}\times \text{1}$ convolution are used to ensure good performance in the case of low depth. As an input, spatial attention map can be selectively fed into the input of each block. In addition, we only need to adjust the parameters of the classifier to easily extend to other datasets without increasing too much computation. A large number of experimental results show that the proposed CEDNN is obviously better than the traditional deep learning methods on DISFA+ and CK+ datasets. After adding spatial attention maps, the result is better than the most advanced AU detection methods. Meanwhile, CEDNN can improve the running speed and reduce the requirements for experimental equipment.

\end{abstract}




\end{frontmatter}

\section{Introduction}
\label{Introduction}
As one of the most comprehensive and objective methods to describe facial expressions, facial action coding system (FACS) has received more and more attention in recent years. It was originally proposed by Ekman in \cite{friesen1978facial} and expanded revised edition in \cite{ekman2002facial}. The revised version specifies 32 facial muscle movement units called Action Unit (AU). These motion descriptors describe the movements of the head, eyes, and other parts, such as cheek lift (AU6), eyebrow rise and fall (AU1, AU2), and lip bite (AU25, AU26). Human facial expression can not be simply described as the movement of some regions, especially when there are compound expressions and micro-expressions. Multiple local areas of the face will alternately or simultaneously appear slight changes in appearance, so it is difficult for classifiers trained based on discriminant methods to describe them. Any facial expression is the result of activation of a group of facial muscles, and each possible facial expression can be described as a combination of several AUs. Compared with the information judgment method, FACS can measure and evaluate facial movement in a quantitative way, so it is more objective and comprehensive.

Over the past few decades, many researchers have contributed to the task of AU detection. The objective of AU detection is to identify and predict AU tags on the facial image. In previous work, landmarks are often used to predefine regions of interest (ROIs) related to AU. Li et al. \cite{7961729, 8253509} used landmarks to generate regions of interest of fixed size and location. Giving higher weights to ROIs enhances the attention of the network. Although this method solves the problem that traditional CNN \cite{7780738,7780969} can not flexibly pay attention to some areas of the image, the size of the generated region of interest is fixed, and some of them can not fully describe the motion of some large areas of AU (such as AU6). In order to solve these problems, Ma et al. \cite{MA201935} proposed a method of using expert prior knowledge to segment human face into several regions. This method still requires landmarks and the scale of the model is large, and the processes of training and prediction are slow. Over both approaches require attention to all predefined areas. However, not all predefined regions of interest are valid for an image. Focusing on invalid regions of interest is not only a waste of computing resources, but also may lead to misjudgment. Shao et al. \cite{shao2018deep,2020JAA} formally applied the concept of attention to AU detection tasks for the first time. However, their detailed attention is limited to the adjacent areas of the initial attention predefined by landmark. It does not highlight the relevant areas that exceed predefined attention. Moreover, the detailed attention learned by the network is independent. When multiple AUs appear at the same time, the ROIs of AUs only carry on the addition operation, which can not effectively capture the relationship between the AUs. Shao et al. \cite{8880512} proposed a method (called ARL) based on attention and relational learning. The combination of channel-based attention and spatial attention learning is used to select and extract local features related to AU. Then pixel-level relational learning is used to extract spatial attention. The number of space-channels attention and pixel-level relationship learning module is the same as that of total AUs. It means that the network size will increase significantly with the increase in the number of AUs, which requires high computing power for experimental equipment.

Using landmarks result in the limitation of ROIs. The large scale of the network places a greater demand on experimental equipment. In order to solve these problems, this paper proposes a computational efficient deep neural network (CEDNN) which can effectively use spatial attention maps and channel attention to automatically detect facial AUs. First of all, we propose a spatial attention map based on the difference image. The difference between the neutral face and the action face is used to determine the region of interest. So it does not depend on the landmarks related to AU. The positions of multiple AUs can be focused in one attention map instead of learning each AU independently, which is more conducive to learning the relationship between AUs. Moreover, the spatial attention maps we generate are all pixel-based, which is similar to the idea of learning pixel-level attention explored in \cite{8880512}. In depth model training, we try to enhance the network's attention to the regions of interest related to AU by adding spatial attention to each block. After that, we propose a computationally efficient neural network structure. It improves the performance of the model by adding a skip connection (similar to ResNet \cite{he2016deep} and DenseNet \cite{huang2017densely}) to the basic module of IGC \cite{8237731}. The reason is that lower-level spatial information and higher-level semantic information can be fed to the CNN simultaneously. And the added $\text{1}\times \text{1}$ convolution to further merge the features and improve the recognition effect of the network. Moreover, except for the $\text{1}\times \text{1}$ convolution, the other convolution operations in the basic block are group convolution, which greatly reduces the number of parameters and computational complexity. In the experiment, we visualize the feature maps of each layer of the network and find that some filters in the feature map do not learn to pay attention to the relevant areas of AU, but pay more attention to the whole image. The idea of channel attention is to enhance the weight of the filter related to the target task and reduce that of the filter which is not related to task. For this reason, we introduce the squeeze-and-excitation (SE) module \cite{8701503} as channel attention into our network structure. The whole recognition framework is end-to-end. The output layer of the network is similar to the classification task, which can be adjusted directly without increasing too much computation.

To sum up, the main contribution of this paper has three aspects:

1) We propose a method to generate a spatial attention map based on difference image, which does not depend on the landmarks related to AU and can simultaneously pay attention to multiple AUs at the pixel level.

2) We propose an end-to-end computationally efficient deep neural network, which dramatically reduces the size of the network and speeds up the computing speed of the network.

3) We have conducted extensive experiments on two benchmark datasets. The experimental results show that the proposed network framework with spatial attention maps is superior to the most advanced methods in AU detection tasks.

\section{Related work}
\label{Related work}
In order to extract effective facial features for AU detection, researchers have tried many methods based on geometric features. Facial landmarks have strong robustness in describing geometric changes. Many traditional research methods are based on facial landmarks or texture features near facial landmarks. Fabian et al. \cite{7780969} proposed a method of AU detection based on geometric changes and local texture. Zhao et al. \cite{7471506} proposed a facial AU detection method (called JPML), which considers both patch learning and multi-tag learning. After that, Zhao et al. \cite{7780738} proposed a deep region multi-label learning (DRML) network that can be trained end-to-end. The identification framework divides the response map evenly into $8\times 8$ regions in the region layer and updates the weight in each patch independently. Song et al. \cite{7163081} proposed a Bayesian model, which uses Bayesian compressed sensing to model the sparsity and co-occurrence of AUs. They use PHOG to extract the features of the image and feed them into the Bayesian model. Although Bayesian models can learn certain AU relationships, the features extracted by using PHOG may not describe facial AU features very well. Wang et al. \cite{8006290} used a three-tier Restricted Boltzmann Machine (RBM) to capture the global relationship between all AUs. Then they put forward a method (called FRBM) which can capture feature relationship and tag relationship at the same time. The above two methods take the difference between the landmark of the peak frame and the neutral frame as the feature and send it into the RBM \cite{6751522}. Although landmark can express some AU changes, the texture changes near the landmarks seem to describe the motion of AU better. Moreover, this method is easy to fail when there is a small change in AU. Eleftheriadis et al. \cite{7410789} simultaneously used facial landmarks and appearance features extracted by local binary pattern (LBP) for AU detection.

Many researchers have also proposed a method to divide the whole face into several regions of interest. Jaiswal \cite{7477625} used facial landmarks to calculate binary image masks to encode shape features. They connect the landmarks that define the image area into a polygonal area in order. The internal elements of them are set to 1, and the others set 0 to get a binary mask image. The image area and the mask area are sent into the network as two input streams. Li et al. \cite{8100199} designed an ROI clipping network to learn the regions of $3\times 3$  around 20 AU centers determined by landmark. Each sub-region has a local convolution neural network. These convolution kernels are independent of each other. Secondly, all the features are connected and sent into the full connection layer to learn the relationship of AUs and obtain the global features across sub-regions for AU detection. Wang et al. \cite{WANG2019130} constructed a network using local convolution and residual units and proposed a slope loss function for multi-label learning. Ma et al. \cite{MA201935} proposed a method to segment human face regions into several regions using expert prior knowledge. They no longer use the method of finding the center of AU but mark the areas where multiple AUs or single AU are located according to the prior knowledge of experts. The bounding box of each region is sent into the target detection network to learn the features of each region separately, and the ``bit-wise OR" operator is used to obtain the final AU detection results of the image-level. Liu et al. \cite{liu2020relation} proposed an AU detection method based on graph convolution network (GCN) for AU relation modeling. Firstly, several regions related to AU are obtained according to the AU center and then sent to the automatic encoders (AEs) to get the potential vectors related to AU. After that, each potential vector is fed back to the GCN as a node. The connection mode of the AUs is determined according to the relationship between the GCN. Finally, the features updated by GCN are spliced to achieve AU detection. Corneanu et al. \cite{corneanu2018deep} proposed a deep structural inference network (DSIN). It uses CNN to extract features of the entire image and patches separately. Then the integrated features are sent to a set of interconnected recursive structure inference units, and the information is transmitted iteratively to infer the structure between AUs. All the above methods depend on the accuracy of landmark detection, and the definition of the AU center is very complicated. Many of these methods learn the features of each AU separately and only learn the relationship between AUs through joint feature learning. This method often loses the spatial relationship of AUs.

Recently, there has been some work on the use of spatial attention mechanism for AU detection. Li et al. \cite{7961729,8253509} proposed an enhanced clipping network (called EAC-NET). They still use the AU center to generate the attention area map of each AU and add these attention area maps to the network to enhance and crop the regions of interest of the AU. The way of generating the attention area map through the AU center is the same, so the predefined ROI of each AU has a fixed size and a fixed attention distribution. Shao et al. \cite{shao2018deep,2020JAA} proposed an adaptive attention learning module (JAA-Net and an improved version J\^{A}A-Net) for joint face AU detection and face alignment tasks. The landmark determines the area of the initial predefined attention. After that, the refined attention map is limited to the adjacent areas of the initial predefined attention. No attention is paid to the relevant areas that exceed the predefined attention map. Moreover, the predefined attention map and refined attention map of a certain AU learned by the network are independent. Shao et al. \cite{8880512} proposed a method (called ARL) based on attention and relationship learning. The features obtained by hierarchical multi-scale regional network learning are sent into the channel attention learning module which is similar to the SE module. The spatial attention learning module composed of two convolution layers is used to select and extract the local features related to AU. After that, the features obtained by the spatial attention module are sent to the single-channel convolution layer and the sigmoid function to process the spatial attention weight. Finally, pixel-level relation learning is used to extract further spatial attention. ARL has the same attention learning and relationship learning structure for each AU. This means that the computational complexity is largely determined by the total number of target AU.
\section{Proposed method}
\label{Proposed method}

\subsection{Overview}
\label{overview}
\begin{figure}[ht]
\centering
\includegraphics[width=13cm]{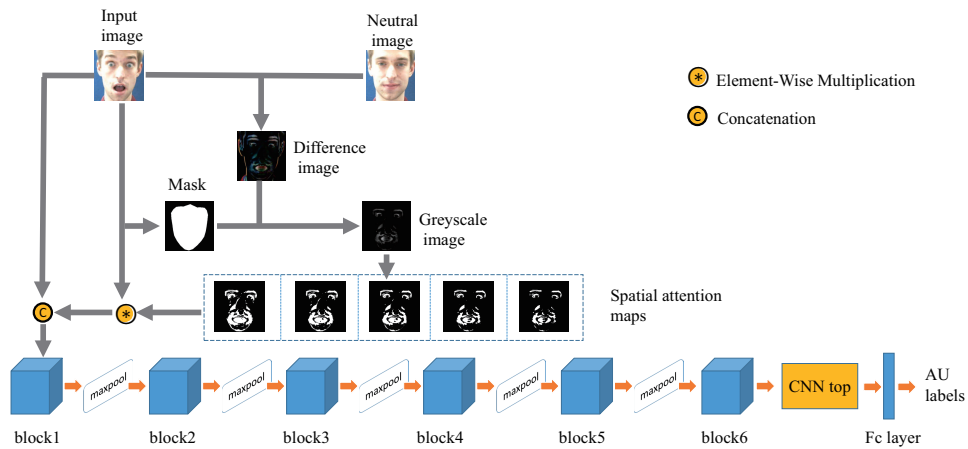}
\caption{Overview of our framework}
\label{fig1}
\end{figure}
We review the generation methods of attention maps in previous AU detection tasks and propose a method of using facial difference images as spatial attention. The whole process consists of three steps. First, the face needs to be aligned before generating difference image. Secondly, a mask is generated according to the coordinates of the landmarks and the connection rules. The mask and the difference image are used to obtain an image containing only the face via a ``bit-wise AND'' operator. Finally, through the binarization process, five binarized images with different thresholds are obtained. Then these five binarized images are sent to our network as spatial attention maps. Classical networks such as VGG \cite{simonyan2014very} and ResNet \cite{he2016deep} are used as backbone structures in previous AU detection tasks \cite{7961729} and \cite{MA201935}. The scale of these networks is very large, and they require high computing power. We use the group convolution with fewer parameters to build the network. After the fusion of the difference images containing only the face and the original input, they are sent to the network to help complete the feature extraction. The final features are sent to the classifier to complete the AU detection task. The specific process is shown in Fig.\ref{fig1}. This section introduces the generation rules of the spatial attention map and the module of the backbone. We also introduce the channel attention module in Section \ref{channel attention}.

\subsection{Spatial attention map}
\label{Spatial Attention Map}
The use of spatial attention map is very common in image-related tasks. In \cite{7961729}, the pixels around the landmarks are used as attention maps. Not all the pixels around the landmarks are essential in an image. For example, when only the muscles around the eyes move, if the area around the mouth is still used as the area of interest, it may confuse the network and fail to find the part paid attention to. The difference image can show the part of the motion. It inspired us to use the difference images between the image with AU and the neutral expression without AU motion as spatial attention maps. Therefore we solve the problem of using certain fixed areas as attention.

\begin{figure}[ht]
  \centering
  \subfigure []{
  \includegraphics[width=2cm]{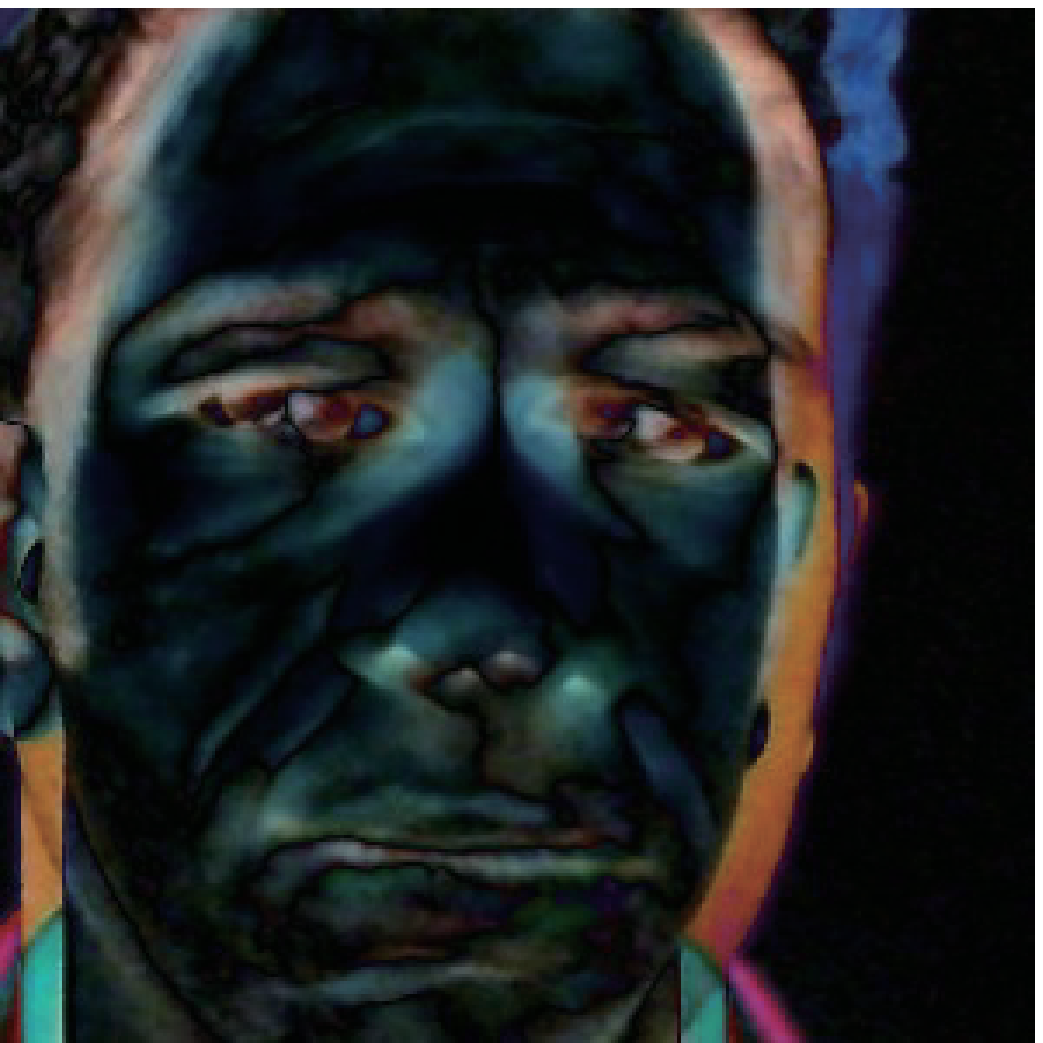}
  }
  \quad
  \subfigure []{
  \includegraphics[width=2cm]{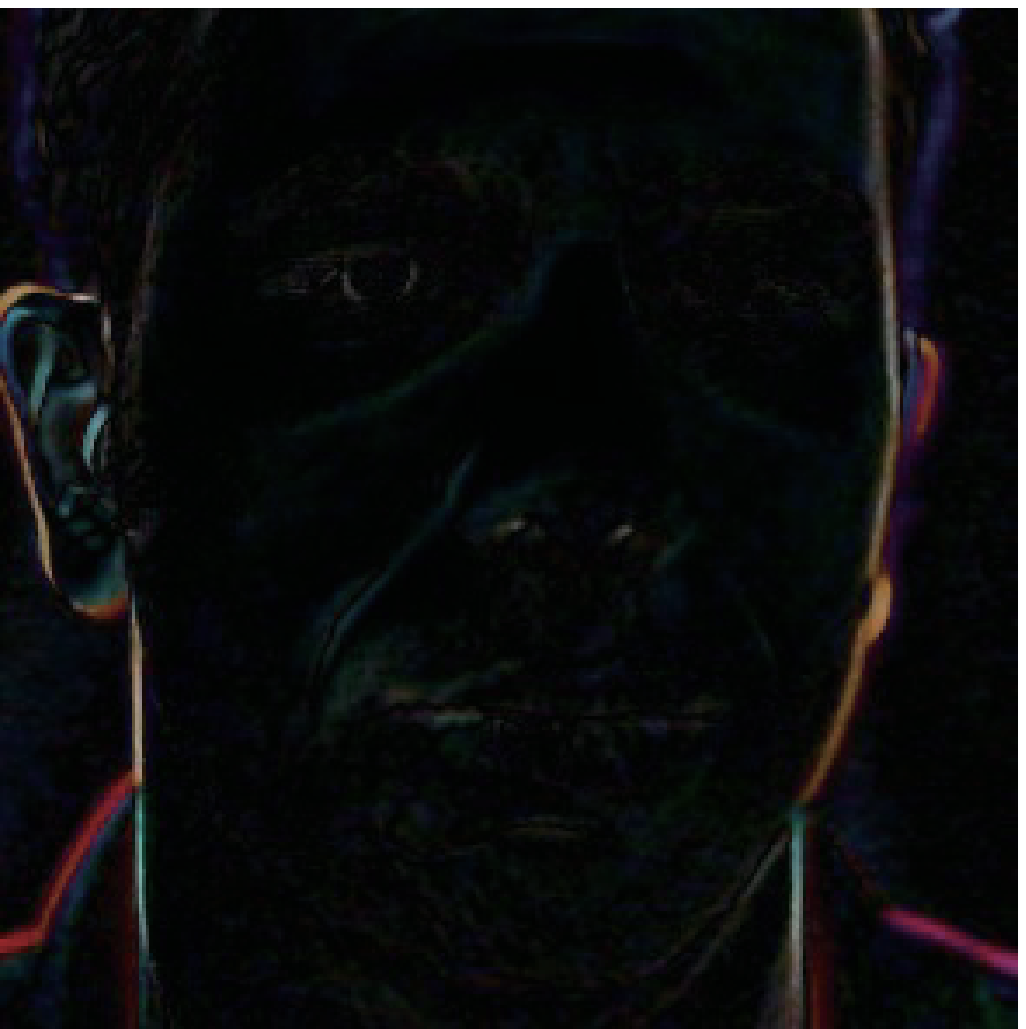}
  }
  \caption{Difference image. (a) before face alignment (b) after alignment}\label{fig2}
\end{figure}

(1) There is inevitably a slight change in the tilt of the subject's head and the position in the camera during the recording of the dataset. Therefore, the first step is to perform face alignment when generating a spatial attention map. We used the dlib \cite{king2009dlib} toolbox to extract five facial landmarks and then carried out the alignment operation. The difference images before and after the alignment operation are displayed in Fig.\ref{fig2}

From Fig. \ref{fig2}, we can see that the two images before alignment are quite different. Their difference images show other differences besides the movement of facial muscles. After the alignment operation, the difference in most areas of the face becomes very small, and only the area where facial muscle movement occurs has obvious changes and high brightness. However, it can be seen from Fig.\ref{fig2}(b). that the brightness of other parts besides the face area is relatively high. If the difference image is directly used as the spatial attention map, too much attention will be paid to information such as facial contours. It will increase the difficulty of network training and affect the final recognition ability.

\begin{figure}[ht]
  \centering
  \subfigure []{
  \includegraphics[width=2cm]{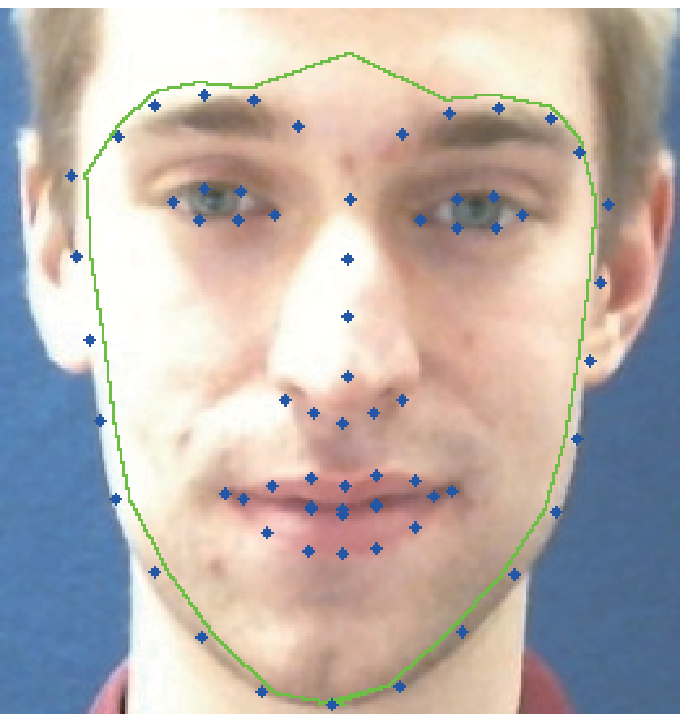}
  }
  \quad
  \subfigure []{
  \includegraphics[width=2cm]{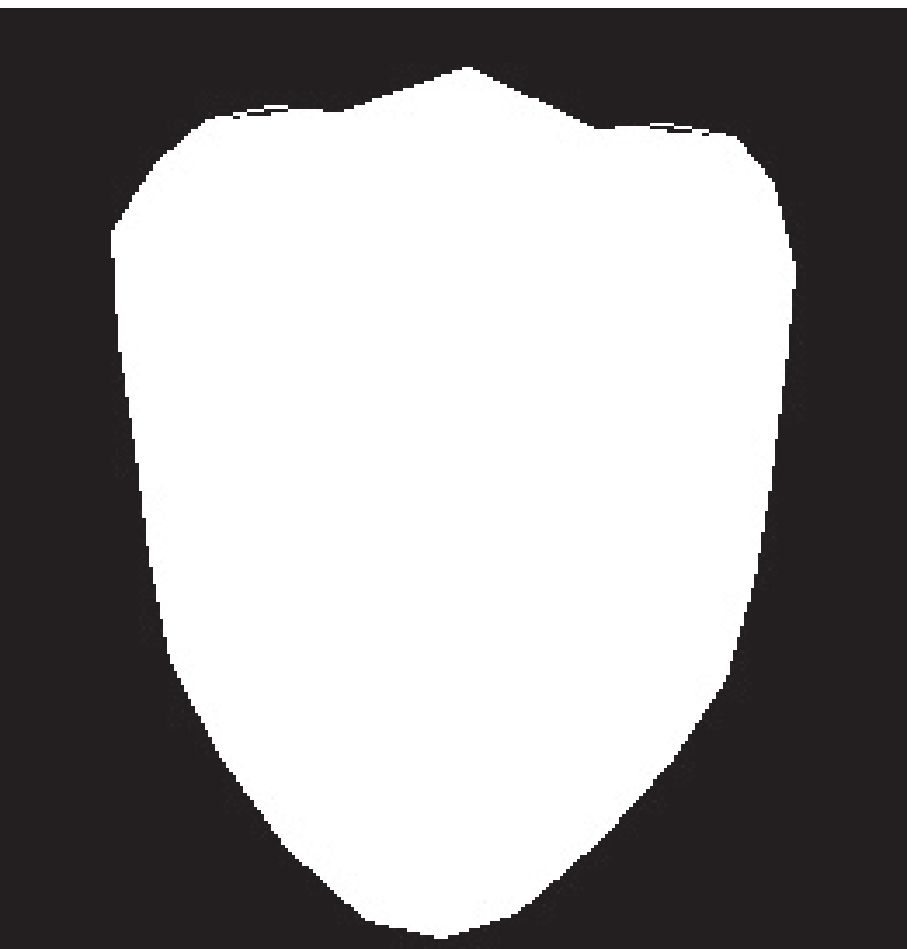}
  }
    \quad
  \subfigure []{
  \includegraphics[width=2cm]{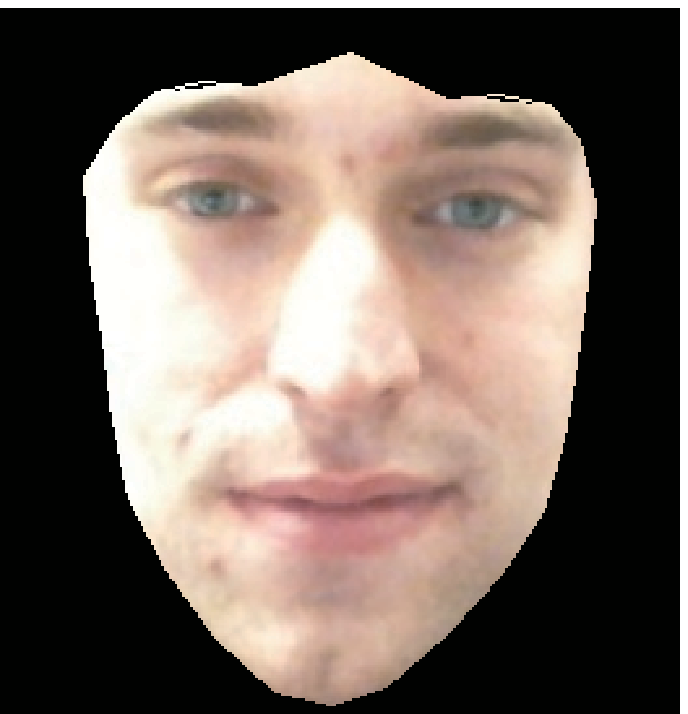}
  }
  \caption{Mask}\label{fig3}
\end{figure}

(2) The second step is to generate a mask. Sixty-four facial landmarks are extracted by using the dlib toolbox. The points representing the facial outline and the points on the eyebrows are used to form a closed area. Then the position of the human face in the image is expressed. Because the aligned differential image in Fig.\ref{fig2} tends to have higher brightness in the facial contour part. When connecting the lines of this closed area, the coordinates of the facial landmarks are not directly used here. The landmark positions of the facial contours are all moved inward by 5 pixels, and the landmarks on the eyebrows are moved up by 5 pixels. In addition, when analyzing the images produced by AU motion, it is found that AU4 is often accompanied by extrusion deformation of the middle muscle of the eyebrow. Instead of using the points between the two eyebrows to connect directly, we take another point. This point is based on the lower inner point of the eyebrows plus the distance from the upper point of the nose. The selection of connection lines is shown in the Fig.\ref{fig3}(a). Connect the calculated landmark positions to generate a mask, as shown in Fig.\ref{fig3}(b). The resulting mask is matched with the original image to get an image containing only the face in the Fig.\ref{fig3}(c).

\begin{figure}[ht]
  \centering
  \subfigure []{
  \includegraphics[width=2cm]{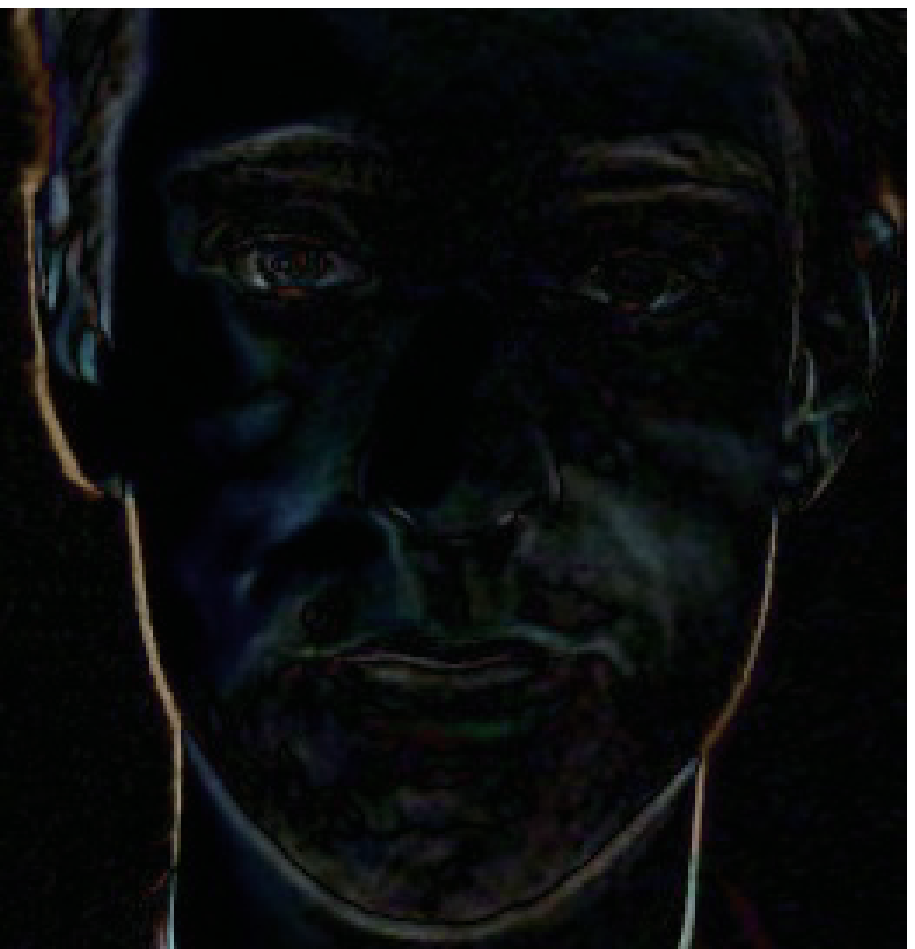}
  }
  \quad
  \subfigure []{
  \includegraphics[width=2cm]{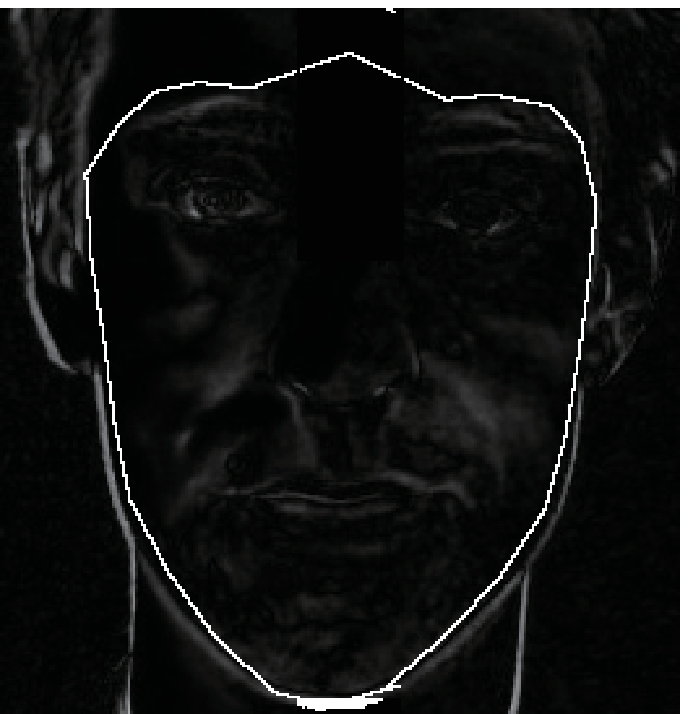}
  }
    \quad
  \subfigure []{
  \includegraphics[width=2cm]{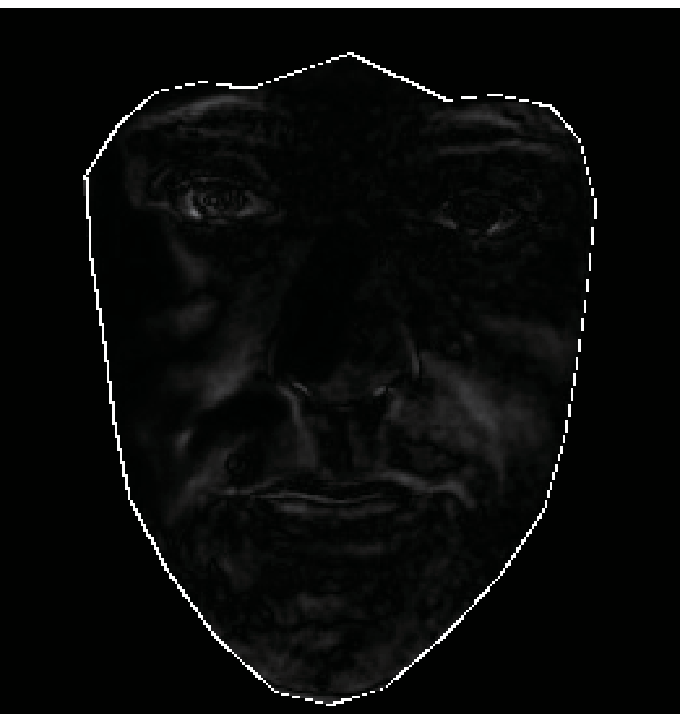}
  }
     \quad
  \subfigure []{
  \includegraphics[width=2cm]{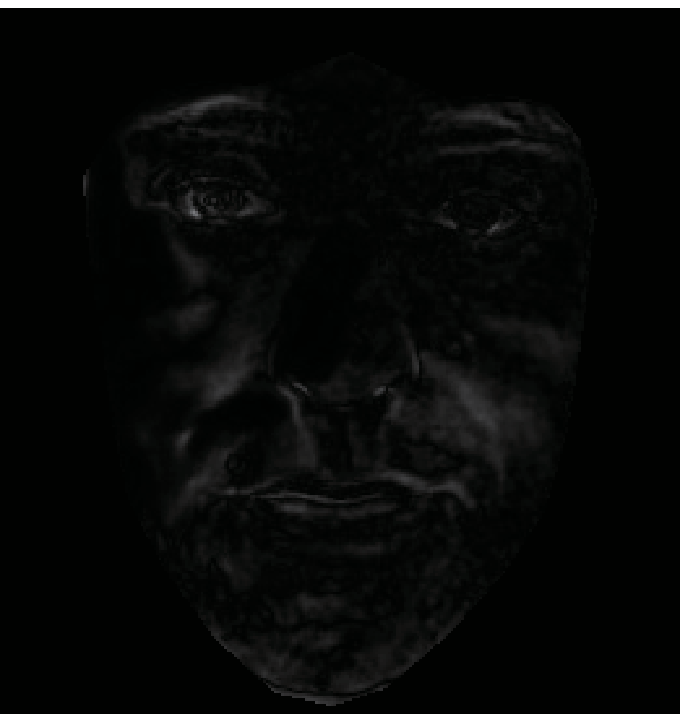}
  }
  \caption{Use the mask to generate difference images containing only the face. (a) The color image after the difference operation. (b) The gray-scaled image with the facial contour connecting line. (c) The image after the  AND operation with the mask. (d) The grayscale image containing only the face after removing the facial contour connecting lines.}\label{fig4}
\end{figure}

After obtaining the mask, the difference image containing only the face is obtained by applying it, as shown in Fig.\ref{fig4}.

\begin{figure}[ht]
  \centering
  \subfigure []{
  \includegraphics[width=2cm]{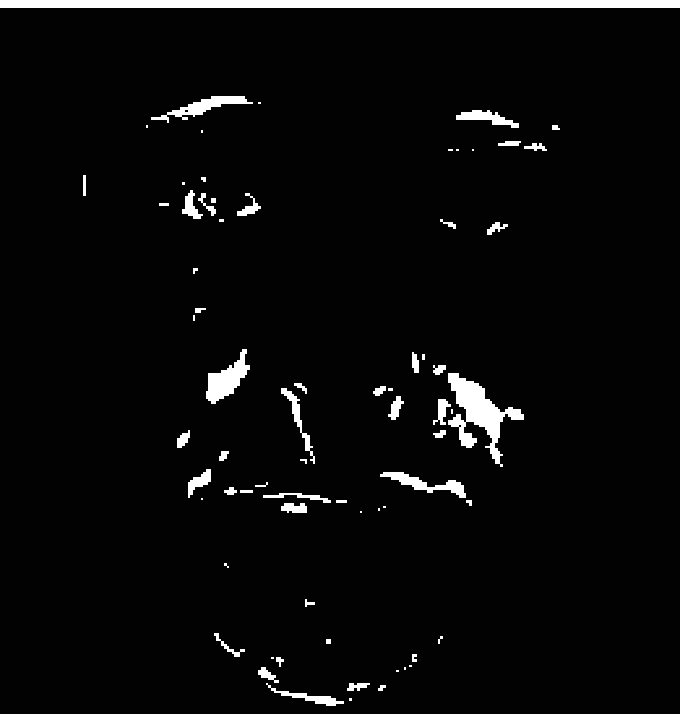}
  }
  \quad
  \subfigure []{
  \includegraphics[width=2cm]{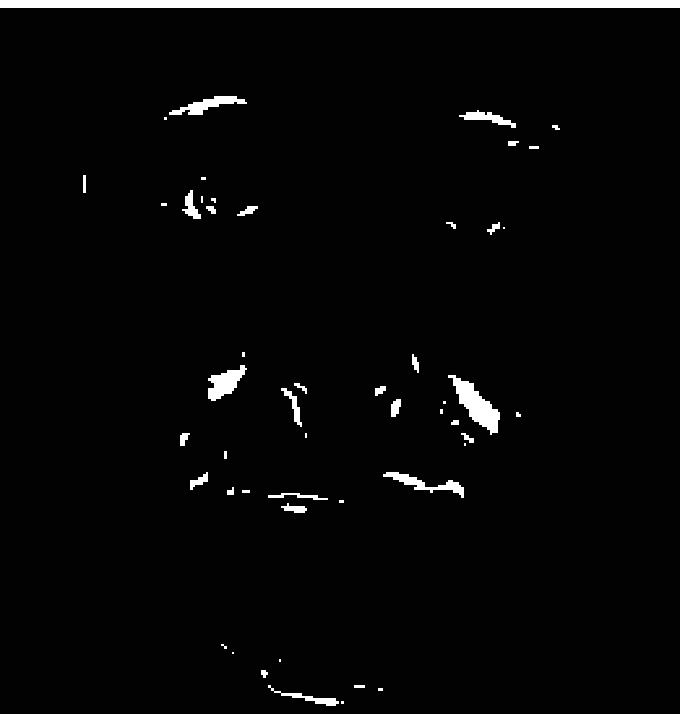}
  }
    \quad
  \subfigure []{
  \includegraphics[width=2cm]{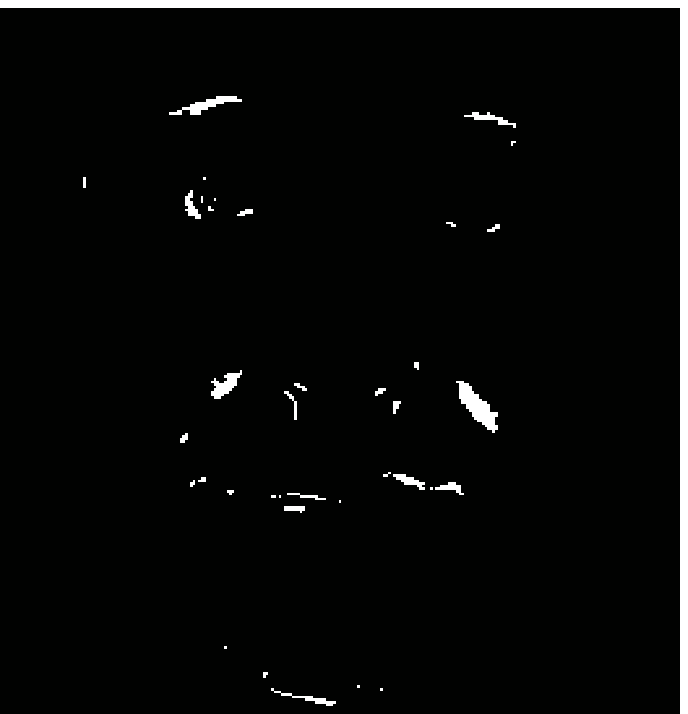}
  }
     \quad
  \subfigure []{
  \includegraphics[width=2cm]{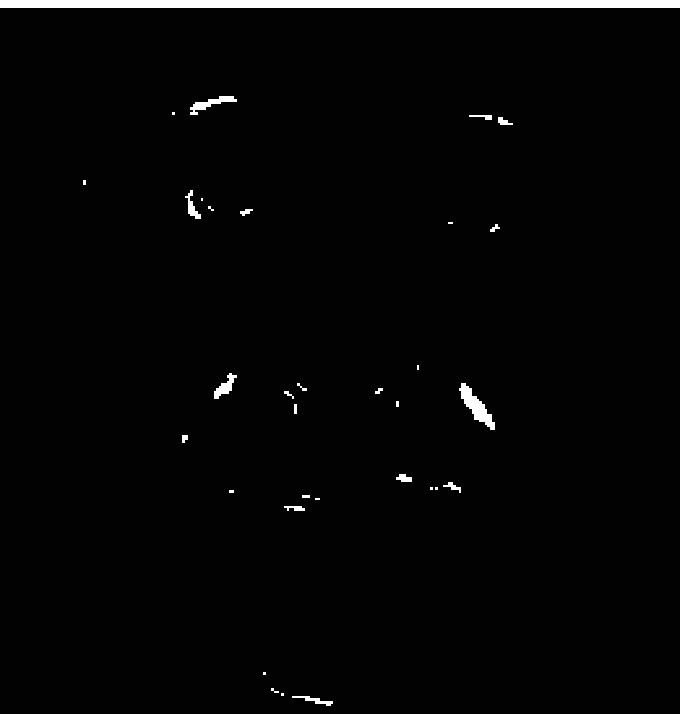}
  }
       \quad
  \subfigure []{
  \includegraphics[width=2cm]{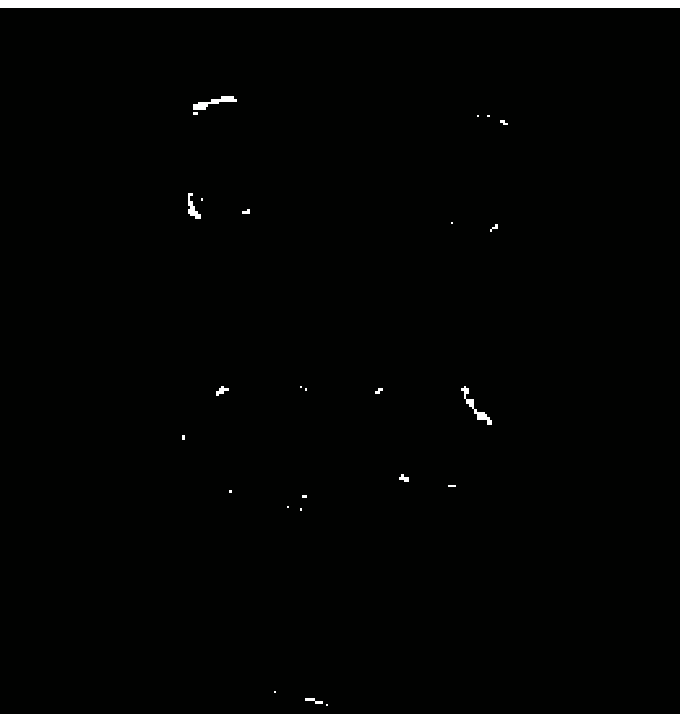}
  }
  \caption{Binary image with different thresholds. (a) 30 (b) 35 (c) 40 (d) 45 (e) 50.}\label{fig5}
\end{figure}

(3) The third step is to generate the binary image of the difference grayscale image. Binarization makes the entire image show a clear black and white effect. The brightness of each pixel in a grayscale image containing only faces is low. The multi-level value of the pixel is involved, and the processing is difficult, so it is not appropriate to send the grayscale image directly to the network as an attention map. Binarizing grayscale images is a common technical method. The collective nature of the binarized image is only related to the position of the point with the pixel value of 0 or 255. It no longer involves the multi-level value of the pixel. All pixels whose grayscale is greater than or equal to the threshold are determined to belong to a specific object, and their grayscale value is 255. Otherwise, these pixels are excluded from the object area, and the grayscale value is 0, which indicates the background or exception object area. Here we set the threshold to $\left\{ 30,35,40,45,50 \right\}$. We have tried to divide each image into several regions and use their mean and variance to calculate the threshold. The variance can reflect the degree of dispersion of the pixel data. When the variance is large, the pixels larger than the mean are set to 255, otherwise set to 0. In most cases, the variance is not large, indicating that the pixel value is close to the mean. The mean is generally relatively small, so there are more elements close to 0 in the region. Therefore, when the variance is small, the area can be set to zero. The images obtained from the experiment also show that it is not appropriate to use the mean as a threshold. Finally, we choose five thresholds of $\left\{ 30,35,40,45,50 \right\}$ to generate a binary image. When the threshold is 30, more pixels are set to 255. It means that binarized images can show more details. As the threshold increases, certain areas are set to zero. Binarized images increasingly tend to show important parts. Some details of the image are sometimes necessary, but sometimes not so important. We finally retained these 5 binary images. In this way, the regions that have common can be focused on, and some details are also preserved. The binary images shown in Fig. \ref{fig5} are obtained by binarizing the difference grayscale images, which only contain the face.

We take the generated binary images as the spatial attention maps. The image needs to be normalized before sent to the neural network. Because there are only two kinds of pixel values, the normalized values are 0 and 1. The normalized attention maps are multiplied with the corresponding elements of the original image and then added to get the input of the network, showing in Adding spatial attention maps module of Fig.\ref{fig6}.

\subsection{Overall network structure}\label{overall network structure}

\begin{figure}[ht]
\centering
\includegraphics[width=13cm]{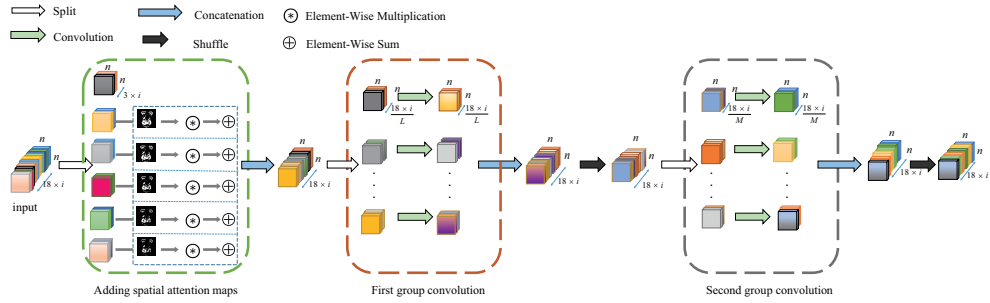}
\caption{The interleaved group convolution module with attention maps.}
\label{fig6}
\end{figure}

The module we build is based on interleaved group convolution (IGC) \cite{8237731} and residual connection \cite{he2016deep} (or dense connection \cite{huang2017densely}). The input channels are divided into several groups, and regular convolution is performed in each group. Therefore, group convolution can be regarded as a regular convolution with a sparse block diagonal convolution kernel, where each block corresponds to a group, and there is no cross-group connection. There are two stages of group convolution in interleaved group convolution, the first group convolution and the second group convolution. Fig.\ref{fig6} shows the partial structure of the interleaved group convolution in each block module, where $n$ represents the size of the input feature map,  $i$ represents the block layer, $L$ represents the number of groups of the first group convolution, and $M$ represents the number of channels in each group of the first group convolution. First of all, the input is divided into several combinations, which are used to deal with spatial correlation. The conventional small convolution kernel $3\times 3$ \cite{simonyan2014very, he2016deep} is usually used to perform convolution in each group. The second group of convolutions is used to fuse the grouping channels of the first group of convolution outputs, and $1\times 1$ convolution (pointwise convolution) is used here.

\textbf{First group convolution} Let $L$ be the number of groups of the first group convolution, and each group contains the same number of channel, which is represented by $M$ here. In IGC, the original input is first sent to a conventional convolution to obtain a multi-channel feature map and then sent to the block of IGC. Therefore, the number of groups can be adjusted by that of regular convolution kernels. We combined five attention maps and an original image to get an 18-channel input. Moreover, we did not use regular convolution but sent the combined input directly into the block. We set $L$ to one of $\left\{ 1,2,3,6,9,18 \right\}$ that can be divisible by 18. Assuming that intra-group convolution is defined as $\mathbf{W}_{ll}^{1}$, then multiple group convolution can be defined as:

\begin{gather}\label{equ1}
\left[ \begin{matrix}
   {{\mathbf{z}}_{1}^{1}}  \\
   {{\mathbf{z}}_{2}^{1}}  \\
   \vdots   \\
   {{\mathbf{z}}_{L}^{1}}  \\
\end{matrix} \right]=\left[ \begin{matrix}
   \mathbf{W}_{11}^{1} & {} & {} & {}  \\
   {} & \mathbf{W}_{22}^{1} & {} & {}  \\
   {} & {} & \ddots  & {}  \\
   {} & {} & {} & \mathbf{W}_{LL}^{1}  \\
\end{matrix} \right]\left[ \begin{matrix}
   {{\mathbf{y}}_{1}}  \\
   {{\mathbf{y}}_{2}}  \\
   \vdots   \\
   {{\mathbf{y}}_{L}}  \\
\end{matrix} \right]\
\end{gather}

where ${{\mathbf{y}}_{l}}$  is a vector of $M\times k$  dimension, and $k$  represents the size of the convolution kernel in the group. For example, when the kernel is $3\times 3$ , $k$  is  9. That is to say,   ${{\mathbf{y}}_{l}}$  corresponds to effective patch in the ${{l}_{th}}$  group. $\mathbf{x}={{\left[ \begin{matrix}
   \mathbf{y}_{1}^{\top } & \mathbf{y}_{2}^{\top } & \cdots  & \mathbf{y}_{L}^{\top }  \\
\end{matrix} \right]}^{\top }}$ represents the input of the first group convolution. ${{\mathbf{z}}^{1}}={{\left[ \begin{matrix}
   \mathbf{z}_{1}^{1\top } & \mathbf{z}_{2}^{1\top } & \cdots  & \mathbf{z}_{L}^{1\top }  \\
\end{matrix} \right]}^{\top }}$
  indicates the output of the first group convolution. It is also the original input of the second group convolution. $\mathbf{W}_{ll}^{1}$ represents the convolution kernel in the ${{l}_{th}}$ group, and is a matrix of size $M\times (M\times k)$.

\textbf{Second group convolution} First, the output of the first group convolution ${{\mathbf{z}}^{1}}={{\left[ \begin{matrix}
   \mathbf{z}_{1}^{1\top } & \mathbf{z}_{2}^{1\top } & \cdots  & \mathbf{z}_{L}^{1\top }  \\
\end{matrix} \right]}^{\top }}$
 is reassembled into $M$ partitions, where $M$ is the number of channel of each group in the first group convolution. According to the shuffle principle in IGC, the ${{m}_{th}}$ channel of each group in the first convolution is selected to form a group. Thus, the channels of each group in the second group convolution come from different first convolution partitions. The number of channel of second group convolution is $L$. The group number of first group convolution remains unchanged. As the number of layers deepens, $M$ is $\left\{ 18\times i,9\times i,6\times i,3\times i,2\times i,1\times i \right\}$, where $i$ is the number of block layers $i\in \left\{ 1,2,3,4,5,6 \right\}$.

\begin{equation}\label{equ2}
\mathbf{z}_{m}^{2}={{\left[ \begin{matrix}
   {{z}_{1m}^{1}} & {{z}_{2m}^{1}} & \cdots  & {{z}_{Lm}^{1}}  \\
\end{matrix} \right]}^{\top }}=\mathbf{P}_{m}^{\top }{{\mathbf{z}}^{1}},\text{ }{{\mathbf{z}}^{2}}={{\mathbf{P}}^{\top }}{{\mathbf{z}}^{1}}
\end{equation}

Among them, $\mathbf{z}_{m}^{2}$ corresponds to the ${{m}_{th}}$ group, $z_{lm}^{1}$ is the ${{m}_{th}}$ element of ${\mathbf{z}_{l}^{1}}$, ${{\mathbf{z}}^{2}}={{\left[ \begin{matrix}
   \mathbf{z}_{1}^{2\top } & \mathbf{z}_{2}^{2\top } & \cdots  & \mathbf{z}_{M}^{2\top }  \\
\end{matrix} \right]}^{\top }}$. $\mathbf{P}$ is the permutation matrix, $\mathbf{P}=\left[ \begin{matrix}
   {{\mathbf{P}}_{1}} & {{\mathbf{P}}_{2}} & \cdots  & {{\mathbf{P}}_{M}}  \\
\end{matrix} \right]$.

The operation performed in the second group convolution is:

\begin{equation}\label{equ3}
{{\mathbf{y}}^{2}}={{\mathbf{W}}^{2}}{{\mathbf{z}}^{2}}
\end{equation}
${{\mathbf{W}}^{2}}$ is a diagonal block matrix, $\mathbf{W}_{mm}^{2}$ on the diagonal represents a convolution kernel of $1\times 1$.

Then through the permutation matrix $\mathbf{P}$, the previous arrangement order of ${{\mathbf{y}}^{2}}$ is restored.
\begin{equation}\label{equ4}
  {{\mathbf{\bar{y}}}^{2}}=\mathbf{P}{{\mathbf{y}}^{2}}
\end{equation}
where ${{\mathbf{y}}^{2}}={{\left[ \begin{matrix}
   \mathbf{y}_{1}^{2\top } & \mathbf{y}_{2}^{2\top } & \cdots  & \mathbf{y}_{M}^{2\top }  \\
\end{matrix} \right]}^{\top }}$. Fig.\ref{fig6} shows the first and second group convolutions mentioned above. We call the above operations a basic block.

\begin{figure}[htp]
\centering
\includegraphics[width=13cm]{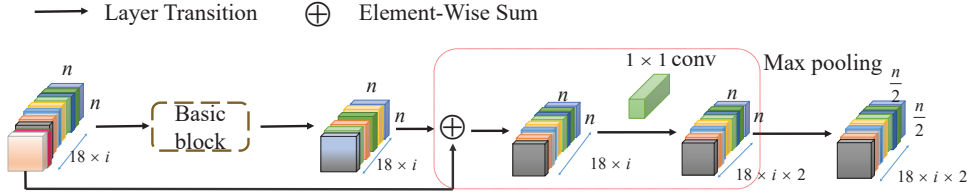}
\caption{Res connection mode}
\label{fig7}
\end{figure}

\begin{figure}[htp]
\centering
\includegraphics[width=13cm]{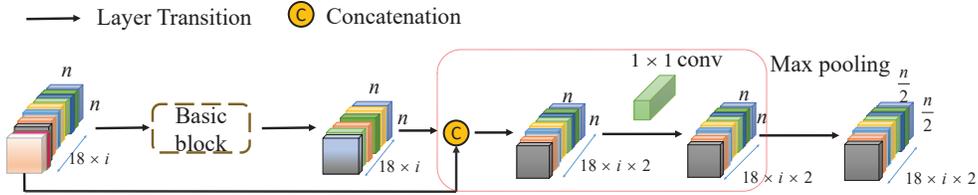}
\caption{Dense connection mode}
\label{fig8}
\end{figure}

\textbf{Connection mode} Our method connects the output of the second group convolution with the input of the entire block module, namely $\mathbf{x}$. \cite{he2016deep,huang2017densely} shows that strengthening information transmission can more effectively use features and improve network performance. We have designed two connection methods, as shown in Fig.\ref{fig7} and Fig.\ref{fig8}. The difference between Fig.\ref{fig7} and Fig.\ref{fig8} lies in part shown in the red box. Firstly, res connection draws lessons from the skip connection of ResNet to add the input of the whole block directly to the output of the second group convolution. The spatial convolution in the first group convolution uses padding to keep the size of the feature map unchanged. The pointwise convolution used in the second group convolution does not affect the size of the feature map. So the output size of the second group convolution ${{\mathbf{\bar{y}}}^{2}}$ is the same as the input of block. They can be added directly.

\begin{equation}\label{equ5}
  {{\mathbf{y}}^{3}}={{\mathbf{\bar{y}}}^{2}}+\mathbf{x}
\end{equation}

The second connection mode, dense connection, draws lessons from the channel merging idea of dense connections in DenseNet and merges the input of the whole block with the output of the second convolution directly. The number of convolution kernels in the second group convolution is the same as the number of input channels, so the number of channel of the second group convolution output is the same as $\mathbf{x}$.
\begin{equation}\label{equ6}
  {{\mathbf{y}}^{3}}=\left[ {{{\mathbf{\bar{y}}}}^{2}},\mathbf{x} \right]
\end{equation}

\bm{$1\times 1$ } \textbf{convolution} After the connection operation, the fusion effect between feature maps may not be excellent. When using res connection, the number of channels in the feature map remains unchanged. The output of the interleaved convolution module has the same number of input channels. If the connected feature map is directly sent to the next block, the number of all feature maps will be the same, and it will always be 18. As the number of layers deepens, increasing the number of convolution kernels can better describe the features. Therefore, a $1\times 1$ convolution operation needs to be added after the connection operation to increase the channel dimension. When using dense connection, connect the input and output directly. There is no fusion between feature maps. The use of a $1\times 1$ convolution kernel ensures that the features are merged twice while the channel dimension is unchanged.

The actions performed by $1\times 1$ convolution is:
\begin{equation}\label{equ7}
  \mathbf{{x}'}={{\mathbf{W}}^{3}}{{\mathbf{y}}^{3}}
\end{equation}
where ${{\mathbf{W}}^{3}}$ represents the convolution kernel of $1\times 1$.

We call the above operation a block. The entire block module can be represented as:
\begin{equation}\label{equ8}
\begin{split}
 & \mathbf{{x}'}={{\mathbf{W}}^{3}}\mathbf{P}{{\mathbf{W}}^{2}}{{\mathbf{P}}^{\top }}{{\mathbf{W}}^{\text{1}}}\mathbf{x}\text{+}{{\mathbf{W}}^{3}}\mathbf{x} \\
 & \text{or }\mathbf{{x}'}={{\mathbf{W}}^{3}}\left[ \mathbf{P}{{\mathbf{W}}^{2}}{{\mathbf{P}}^{\top }}{{\mathbf{W}}^{1}},\mathbf{I} \right]\mathbf{x} \\
\end{split}
\end{equation}

\textbf{Max pooling} Increasing the number of convolution kernels, that is, the channel dimension of the feature map, can obtain more high-dimensional feature information. However, none of the above operations changes the size of the feature map. We need to downsample to expand the receptive field. Our spatial attention map can be added to the input of multiple blocks, so the size of the spatial attention map is also required to be reduced. The convolution operation of the fixed convolution kernel or the average pooling will change the pixel value on the spatial attention map. Using max pooling can reduce the size of the feature map and ensure that the corresponding position pixel on the spatial attention map remains at 1. That is to say, it can not only ensure that the spatial information is not lost but also avoid position movement.

In addition, before feeding the features extracted by the block module to the CNN top module, we first sent the feature map to a $1\times 1$ convolution for channel dimensionality reduction, reducing 1152 channels to 144 channels. Convolution kernel parameters and computational complexity are also reduced. Then features maps are sent to the CNN top module. The CNN top module is composed of 1024 $7\times 7$ large convolution kernels and 1024 $1\times 1$ convolution kernels. Since the size of the input feature map is also $7\times 7$, the operation performed by the large convolution kernel is similar to average pooling. However, the weight of this convolution kernel can be trained, that is, the features can be further extracted. The subsequent $1\times 1$ convolution operation is actually a fully connected layer operation \cite{7478072}. Its operating speed is improved compared to a fully connected network.

\subsection{Channel attention}\label{channel attention}
\begin{figure}[ht]
  \centering
  \subfigure []{
  \includegraphics[width=4cm]{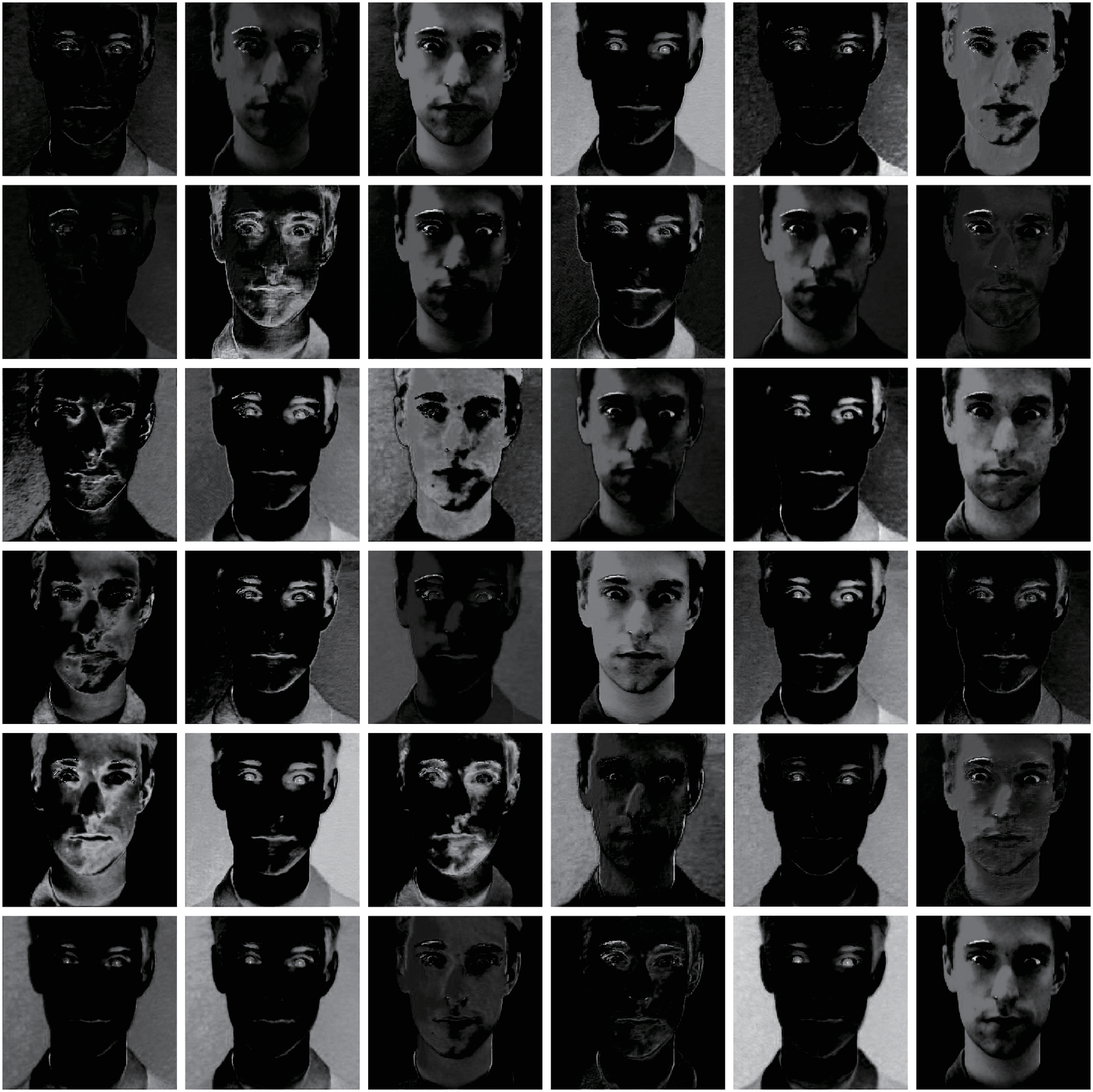}
  }
  \subfigure []{
  \includegraphics[width=4cm]{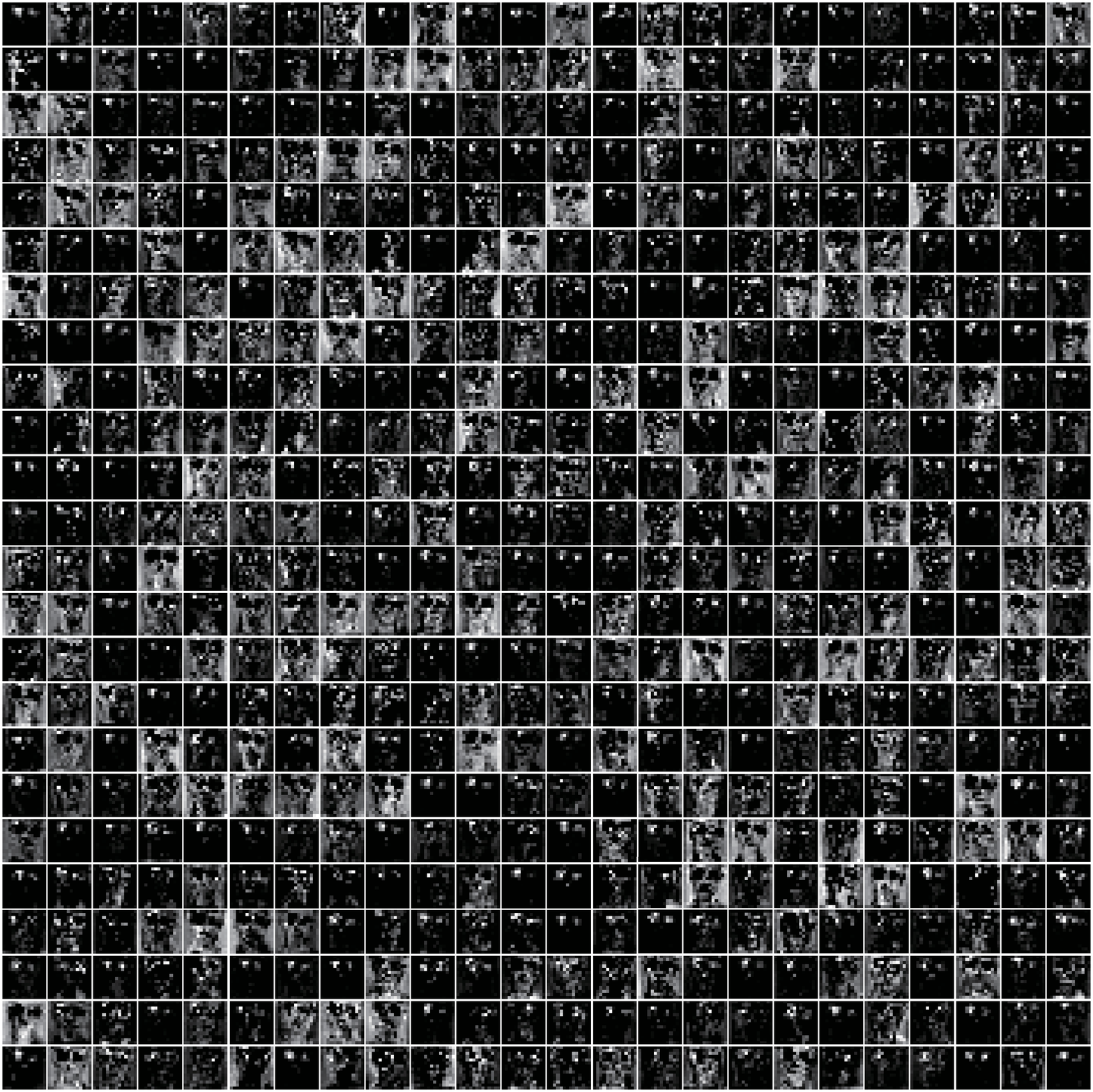}
  }
  \caption{Visualization of features in CEDNN without channel attention module. The feature maps are extracted by block. (a) block1 (b) blcok5}
  \label{fig9}
\end{figure}

Fig.\ref{fig9} shows the visualization of the feature map of two blocks. Fig.\ref{fig9} (a) is the output visualization of block1, Fig.\ref{fig9} (b) is that of block5. It can be seen from Fig.\ref{fig9} (a) that after feature extraction of a module, some feature maps can show the area of interest, such as the feature maps in the first row and fifth column. However, many feature maps do not show the desired area. They still extract the features of the entire image, such as the feature map in the first row and third column of the figure. Corresponding regularity can still be observed in the feature maps of other blocks. Some feature maps can show the corresponding AU area, and some feature maps still focus on the entire image. When the output feature map block5 (Fig.\ref{fig9} (b)) is close to the whole image, it will have a very bad influence on the final classification. Based on the above phenomenon, we have added channel attention to each block. The features of each channel actually represent new features after convolution with different convolution kernels. The contribution of the new $n$ channels to key information must be more or less. If we add a weight to each channel to represent the correlation between that channel and key information, the greater the weight, the higher the correlation. That is, the more we need to pay attention to the channel. The purpose of channel attention is to multiply the values of different channels with different weights, thereby enhancing the attention to key channels. In our network, the SE module \cite{8701503} is used to realize the attention to the channel.

The essence of the SE module is the channel attention module. It learns the weight of each channel through three parts: squeeze, excitation, and attention.

\begin{figure}[htp]
\centering
\includegraphics[width=13cm]{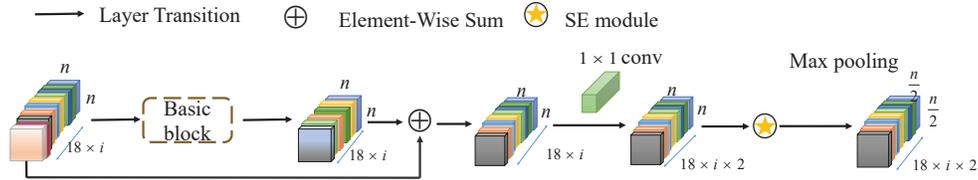}
\caption{The schema of the first se-block module}
\label{fig10}
\end{figure}

\begin{figure}[htp]
\centering
\includegraphics[width=13cm]{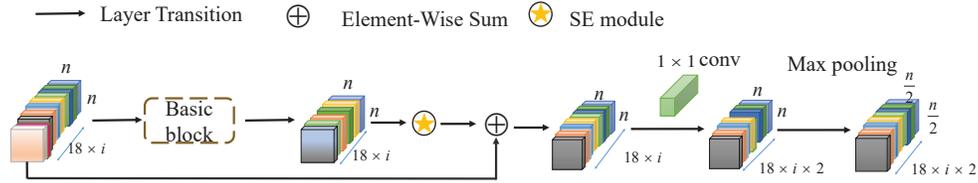}
\caption{The schema of the second se-block module}
\label{fig11}
\end{figure}

The se-module can be integrated into standard architectures such as Inception \cite{ioffe2015batch} and ResNet by insertion after non-linearity following each convolution. Drawing lessons from the two construction methods in \cite{8701503}, the block of CEDNN can be similar to Inception or ResNet with skip connections. Here, we simply convert the convolution operation before the se-module into a complete block (see Fig.\ref{fig10}). By making changes to each of these modules in the architecture, we have obtained the network under the first connection mode, the first se-block module. SE module can also be added to jump-connected modules. Both squeeze and excitation work before summation (or connection) with the identity branch. Similarly, each block in the network is modified according to this addition method to obtain the network with the second connection mode. Fig.\ref{fig11} describes the block in the second connection mode. The same integration method can be extended to the dense-connected block.

\section{Experiments}\label{experiments}
\subsection{Setting}\label{setting}
\subsubsection{Dataset description}\label{Dataset description}
In this section, we evaluate our method on two benchmark AU detection datasets, namely DISFA+ dataset \cite{7789672} and CK+ dataset \cite{lucey2010extended}. Each database is FACS encoded by experienced coders.

(1) DISFA+ contains videos of poses and spontaneous facial expressions from 9 participants. The intensity scores of 12 AUs for all frames are annotated. Each subject watched a 3-minute demonstration, and related equipment recorded their facial expressions on multiple tracks. In each test, the subjects were asked to imitate the full dynamics of facial movements (that is, starting with a neutral face, continuing to the maximum expression intensity, and ending with a neutral face). Each subject was asked to imitate 30 facial movements (single AU or combination of AUs) and 12 facial expressions (for example, surprise, anger, etc.). For DISFA+, a three-fold division is adopted by following previous related work to ensure that subjects are mutually exclusive in the training/testing set. The specific division is shown in Table \ref{tab:tab1}. We used 378 videos (9 subjects $\times $42 videos), which produced $\sim$59,000 valid facial images. Frames with an intensity equal to or greater than 2-level are regarded as positive frames, and other frames are regarded as negative frames.

(2) The CK+ dataset contains 593 records of pose and non-pose sequences in a laboratory environment, which come from 123 subjects. FACS encoding was performed on the last frame (peak expression) in each sequence, which resulted in 593 frames. CK+ contains a variety of AUs, some of which appear unfrequently. Following the use of AU in previous related work \cite{7163081,8006290}, AUs with a frequency of more than 10\% were selected for evaluation. A total of 13 AUs were selected on CK+ to report experimental results. After filtering relevant AUs in CK+, 579 frames are obtained. Table \ref{tab:tab2} shows the number of occurrences of each AU in the CK+ dataset.

\begin{table*}
\centering
\caption{Subjects included in each fold after partitioning the DISFA+ dataset}
\begin{tabular}{cccc} 
\toprule
Fold & \multicolumn{3}{c}{Subjects}   \\

  \midrule
  1 &   SN027&  SN010&  SN007\\
  2 &   SN013&  SN025&  SN003\\
  3 &   SN009&  SN001&  SN004\\
  \bottomrule
  \end{tabular}
  \label{tab:tab1}
\end{table*}

\begin{table}[htbp]
 \centering
 \caption{Number of times that each AU appears in the CK+ dataset }
  \begin{tabular}{cccccccccccccc}
  \toprule
       AU   &    1& 2& 4& 5& 6& 7& 9& 12& 17& 23& 24& 25& 27  \\
  \midrule
  Num &   175&  117& 192& 102& 123& 120& 75& 130& 201& 60& 58& 323& 81\\
  \bottomrule
  \end{tabular}
 \label{tab:tab2}
\end{table}

\subsubsection{Metric}\label{Metric}
Detecting whether AU is active is a multi-label binary classification problem. For binary classification tasks, especially when the samples are unbalanced, F1 scores can better describe the performance of the algorithm \cite{7410789,7284874}. Similar to the previous methods \cite{7961729,8100199,7780738}, frame-based F1 Score (F1-Frame, \%) is reported. The F1 score consists of two components: precision ($p$) and recall ($r$). The precision is called the positive predictive value, which is the fraction of true positive predictions to all positive predictions. The recall is also called sensitivity, which is the score of true positive prediction relative to all ground-truth positive samples. Knowing $p$ and $r$, we can get the F1 score, as shown in Eq.(\ref{equ9}). In our evaluation, we calculated the F1 scores of 13 AUs in CK+ and 12 AUs in DISFA+. The F1 score can be directly used as an indicator to compare the performance of different algorithms on each AU. The overall performance of the algorithm is described by the average F1 scores of all AUs (Avg). In the following sections, we will omit \% in all results for simplicity.

\begin{equation}\label{equ9}
  F1=\frac{2p\times r}{p+r}
\end{equation}

\subsubsection{Implementation details}\label{Implementation Details}
The face is detected and cropped to obtain an $n\times n\times 3$ color image, where $n$ represents the pixels occupied by the face in the original image. The face image is resized to $224\times 224$. The input face image should correspond to the position of the generated spatial attention map. Except for the resize operation, no other data enhancement operations are performed. All training processes do not use pre-trained weights. The model is initialized with a learning rate of 0.01. The learning rate of CK+ dataset is reduced by a factor 0.1 after every 20 epochs, and that of DISFA+ dataset is reduced by 0.1 times every 5 epochs. In all experiments, we chose momentum stochastic gradient descent to train CK+ for 30 epochs (DISFA+ for 20 epochs) and set momentum to 0.9, weight attenuation to 0.0005, and mini-batch size to 32. All implementations are based on pytorch, and training and testing are done on NVIDIA GeForce RTX 2080Ti.

\subsubsection{Compared methods}\label{Compared Methods}
Under the same settings described in Section \ref{Implementation Details}, we compared our method with the state-of-the-art AU detection methods. When evaluating on the DISFA+ dataset, we compared CEDNN with DRML \cite{7780738}, AU R-CNN \cite{MA201935} and J\^{A}A-Net \cite{2020JAA}. J\^{A}A-Net is the latest research result and an improved version of pioneering work based on attention. On the CK+ dataset, it is compared with the recent related work BGCS \cite{7163081}, HRBM \cite{6751522}, JPML \cite{7471506}, DSCMR \cite{WANG2019130}.

\subsection{Conventional CNN versus CEDNN}\label{conventional CNN versus CEDNN}
CEDNN is proposed for AU detection in Section \ref{Proposed method}. The architecture is shown in Table \ref{tab:tab3}. Therefore, our first experiment aims to determine whether it can perform better than the baseline conventional CNN. All experiments use the entire face RGB image for learning. We believe that the recognition ability can be improved by adding spatial attention maps, and the number of parameters can be greatly reduced by using group convolution, and the requirements for hardware computing power can also be reduced. We train classic CNNs and CEDNN on CK+ and DISFA+ datasets for comparison. The classic CNNs include ResNet18, DenseNet121 and IGC-L24M2 . 13 AUs are used in CK+, and 12 AUs are used in DISFA+. All models use the Sigmoid cross-entropy loss function. In each iteration, we randomly select 32 images to form a small batch and initialize the learning rate to 0.01.

\begin{table}[htbp]
 \centering
 \caption{The architecture of CEDNN (such as the L=6, M=3 in the block). The third column indicates the number of branches and channels of the first group convolution. The fourth column represents the number of branches and channels of the second group convolution. The fifth column illustrates whether to add spatial attention maps.}

  \begin{tabular}{ccccc}
  \toprule
          Output size&    Layer name& \tabincell{c}{branches \\ /channels}& \tabincell{c}{branches \\/ channels}& attention maps\\
  \midrule
  $224\times 224$ &   block1&  6/3& 3/6& Yes\\
  $112\times 112$ &   Max-pooling&  -& -& -\\
  $112\times 112$ &   block2&  6/6& 6/6& Yes\\
  $56\times 56$ &   Max-pooling&  -& -& -\\
  $56\times 56$ &   block3&  6/12& 12/6& No\\
  $28\times 28$ &   Max-pooling&  -& -& -\\
  $28\times 28$ &   block4&  6/24& 24/6& No\\
  $14\times 14$ &   Max-pooling&  -& -& -\\
  $14\times 14$ &   block5&  6/48& 48/6& No\\
  $7\times 7$ &   Max-pooling&  -& -& -\\
  $7\times 7$ &   block6&  6/96& 96/6& No\\
  $1\times 1$ &   CNN top&  -& -& -\\
  -& Fc& -& -& -\\
  \bottomrule
  \end{tabular}

 \label{tab:tab3}
\end{table}

\begin{table}[htbp]
\centering
\caption{F1 score result comparison with conventional CNNs on DISFA+. Results of CEDNN without attention maps are
reported using F1 score on 3-fold protocol. Bracketed bold numbers indicate the best score; bold numbers indicate the second best.}
\resizebox{\textwidth}{!}
{
\begin{tabular}{cccccccccccccccc}
\toprule
\multirow{2}{*}{AU} &\multicolumn{3}{c}{CNN} & \multicolumn{6}{c}{Res connection}&\multicolumn{6}{c}{Dense connection}\\
\cmidrule(r){2-4}\cmidrule(r){5-10}\cmidrule(r){11-16}
& ResNet18&DenseNet121 &IGC &L1M18 & L2M9 & L3M6 & L6M3 & L9M2 & L18M1 & L1M18 & L2M9&L3M6 &L6M3& L9M2&L18M1 \\
\midrule

1& 71.9	&67.5	&73.9	&79.4	&74.7	&75.6	&79.7	&78.2	&\boldmath $[80.5]$ \unboldmath	&76.8& 78.2& \boldmath $80.1$ \unboldmath& 79.3& 76.6& 79.6\\

2& 73.5& 70.2& 76.7& \boldmath $[79.7]$ \unboldmath & 72.8& 77.2& 78.2& 75.7& 74.8& 78.0& 77.9& 76.9& 74.0&79.2& \boldmath $79.5$ \unboldmath \\

4& 65.8& 67.8& 81.5& 81.7& 79.0& 74.1& 81.9& 79.2& \boldmath $82.1$ \unboldmath& \boldmath $[82.7]$ \unboldmath& 80.5& 79.0& 78.2& 81.3& 78.2 \\

5& 58.0& 65.1& 55.9& 64.3& \boldmath $74.7$ \unboldmath& 72.8& 68.2& 62.7& 69.4& 67.2& 68.0& 60.8& 65.5& 63.4& \boldmath $[75.6]$ \unboldmath \\

6& 64.7& 61.7& 74.4& 80.5& 80.2& 78.6& 76.3& 77.0& \boldmath $[84.2]$ \unboldmath& 81.0& \boldmath $[84.1]$ \unboldmath& 80.6& 78.1& 73.7& 78.5 \\

9& 45.3& 45.5& 56.8& 65.4& \boldmath $72.8$ \unboldmath& 70.1& 68.9& 69.9& 67.1& 71.8& \boldmath $[77.1]$ \unboldmath& 66.8& 71.4& 61.0& 66.5 \\

12& 63.4& 57.6& 70.8& 69.4& 69.9& 67.1& 64.0& 62.3& \boldmath $[71.3]$ \unboldmath& 70.8& 66.5& 68.8& 66.9& 65.6& \boldmath $[71.3]$ \unboldmath \\

15& 7.8& 39.5& 56.1& 55.0& 55.5& 54.2& 51.7& 46.2& 51.9& 54.8&  \boldmath $57.6$ \unboldmath& 53.7& 53.0& \boldmath $[58.8]$ \unboldmath& 55.0 \\

17& 28.8& 42.5& 56.2& 53.4& 50.6& 59.3& 53.6& 56.0& \boldmath $[60.7]$ \unboldmath&56.7& \boldmath $59.6$ \unboldmath& 57.4& 52.8& 55.5& 51.1 \\

20& 28.7& 32.0& 31.1& \boldmath $41.3$ \unboldmath& 33.3& 35.6& \boldmath $[42.4]$ \unboldmath&27.5& 37.4& 36.2& 35.6& 27.6& 28.8&32.2& 39.2 \\

25& 77.3& 81.8& 82.4& 79.9& 84.7& 88.9& 88.1& \boldmath $89.6$ \unboldmath& 85.2& 85.8& 86.7& \boldmath $[90.4]$ \unboldmath& 89.2& 87.2& 84.0 \\

26& 58.4& 65.0& 70.6& 67.4& 73.2& 73.6& 73.4& \boldmath $[75.4]$ \unboldmath& 69.5& 73.7& 71.5& \boldmath $74.3$ \unboldmath& 70.5& 68.5& 74.2 \\

Avg& 53.6& 58.0& 65.5& 68.1& 68.4& 68.9& 68.9& 66.7& 69.5& \boldmath $69.6$ \unboldmath& \boldmath $[70.3]$ \unboldmath& 68.0& 67.3& 66.9& 69.4 \\
\bottomrule

\end{tabular}
}
\label{tab:tab4}
\end{table}

\begin{table}[htbp]
\centering
\caption{F1 score result comparison with conventional CNNs on CK+. Results of CEDNN without attention maps are
reported using F1 score on 3-fold protocol. Bracketed bold numbers indicate the best score; bold numbers indicate the second best.}
\resizebox{\textwidth}{!}
{
\begin{tabular}{cccccccccccccccc}
\toprule
\multirow{2}{*}{AU} &\multicolumn{3}{c}{CNN} & \multicolumn{6}{c}{Res connection}&\multicolumn{6}{c}{Dense connection}\\
\cmidrule(r){2-4}\cmidrule(r){5-10}\cmidrule(r){11-16}
& ResNet18&DenseNet121 &IGC &L1M18 & L2M9 & L3M6 & L6M3 & L9M2 & L18M1 & L1M18 & L2M9&L3M6 &L6M3& L9M2&L18M1 \\
\midrule

1& 74.9& 67.1& 71.5& 79.9& 80.5& 81.2& \boldmath $[82.7]$ \unboldmath& 79.3& 79.9& 77.2& 80.6& \boldmath $82.5$ \unboldmath& 78.0& 81.8& 80.4\\

2& 85.4& 76.7& 81.5& \boldmath $[88.0]$ \unboldmath& 87.2& 86.2& 84.5& 86.1& 85.8& 84.4& 86.6& 87.2& 87.6& 87.3& \boldmath $87.7$ \unboldmath \\

4& 64.9& 44.7& 69.4& \boldmath $[78.8]$ \unboldmath&75.0& 78.0& 71.9& 76.9& 74.0&  \boldmath $78.0$ \unboldmath& 74.9& 75.0& 73.5& 76.3& 74.5 \\

5& 70.7& 70.5& 70.8& 73.7& 74.7& 74.9& 71.9& 69.5& 74.5& 70.7& 73.8& 72.1& 75.3& \boldmath $76.4$ \unboldmath& \boldmath $[77.5]$ \unboldmath \\

6& 62.1& 25.0& 63.4& \boldmath $[71.6]$ \unboldmath&67.6& 66.7& 71.0& 67.8&  \boldmath $71.4$ \unboldmath& 68.4& 69.0& 69.3& 70.2& 68.1& 69.8 \\

7& 53.5& 26.6& 50.5& \boldmath $57.8$ \unboldmath& 55.4 & \boldmath $[57.9]$ \unboldmath& 55.9& 50.2& 56.2& 56.0& 51.6& 56.0& 53.0& 56.9& 55.0 \\

9& 53.7& 5.2& 79.7& 84.9& \boldmath $86.9$ \unboldmath& 82.6& 81.7& 83.9& 86.5& 85.5& \boldmath $[88.9]$ \unboldmath & 86.7& 84.9& 81.2& 86.3\\

12& 75.9& 30.6& 83.9& 84.2& 81.7& 85.0& 84.6& 83.6& 86.5& 83.4&  \boldmath $[87.8]$ \unboldmath&84.0& 85.1& 85.9& \boldmath $87.2$ \unboldmath \\

17& 77.7& 57.6& 79.6& 85.6& 85.2& 84.9& 85.3& 84.6& 85.8& 82.1& 85.1& \boldmath $86.0$ \unboldmath&82.4& \boldmath $[86.4]$ \unboldmath&85.1  \\

23& 0& 0& 21.6& 27.9& 34.9& \boldmath $[36.2]$ \unboldmath& 25.3& 23.3& 24.1& 35.7&35.7& 30.2& 29.3& \boldmath $35.8$ \unboldmath& 28.6 \\

24& 6.0& 0& 24.7& 30.8& 32.9& \boldmath $40.5$ \unboldmath& 30.2& 37.8& 34.5& 39.6& 39.5& 34.5 & \boldmath $[40.9]$ \unboldmath& 33.7& 24.4  \\

25& 88.6& 81.1& 89.3& 92.1& 93.2 & \boldmath $[93.4]$ \unboldmath& 92.8& 92.9& 92.4& 92.8& \boldmath $93.3$ \unboldmath& 92.7& 90.5& 92.0& 91.7 \\

27& 73.3& 66.2& 84.4& 84.8& 88.6& 85.5& 86.2& 86.8& 86.3& 87.9& \boldmath $90.0$ \unboldmath& 88.5& 83.4 & \boldmath $[90.1]$ \unboldmath&88.3  \\

Avg& 60.5& 42.4& 66.9& 72.3& 72.6& \boldmath $73.3$ \unboldmath&71.1& 71.0& 72.1& 72.4 &\boldmath $[73.6]$ \unboldmath& 72.7& 71.9& 73.2& 72.0 \\
\bottomrule

\end{tabular}
}
\label{tab:tab5}
\end{table}

\begin{figure}[htp]
\centering
\includegraphics[width=13cm]{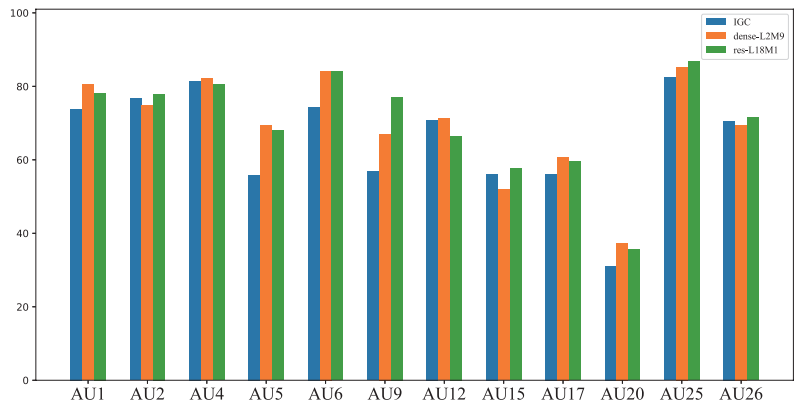}
\caption{The F1 scores of IGC, res-L18M1 and dense-L2M9 for each AU on the DISFA+ dataset.}
\label{fig12}
\end{figure}

\begin{figure}[htp]
\centering
\includegraphics[width=13cm]{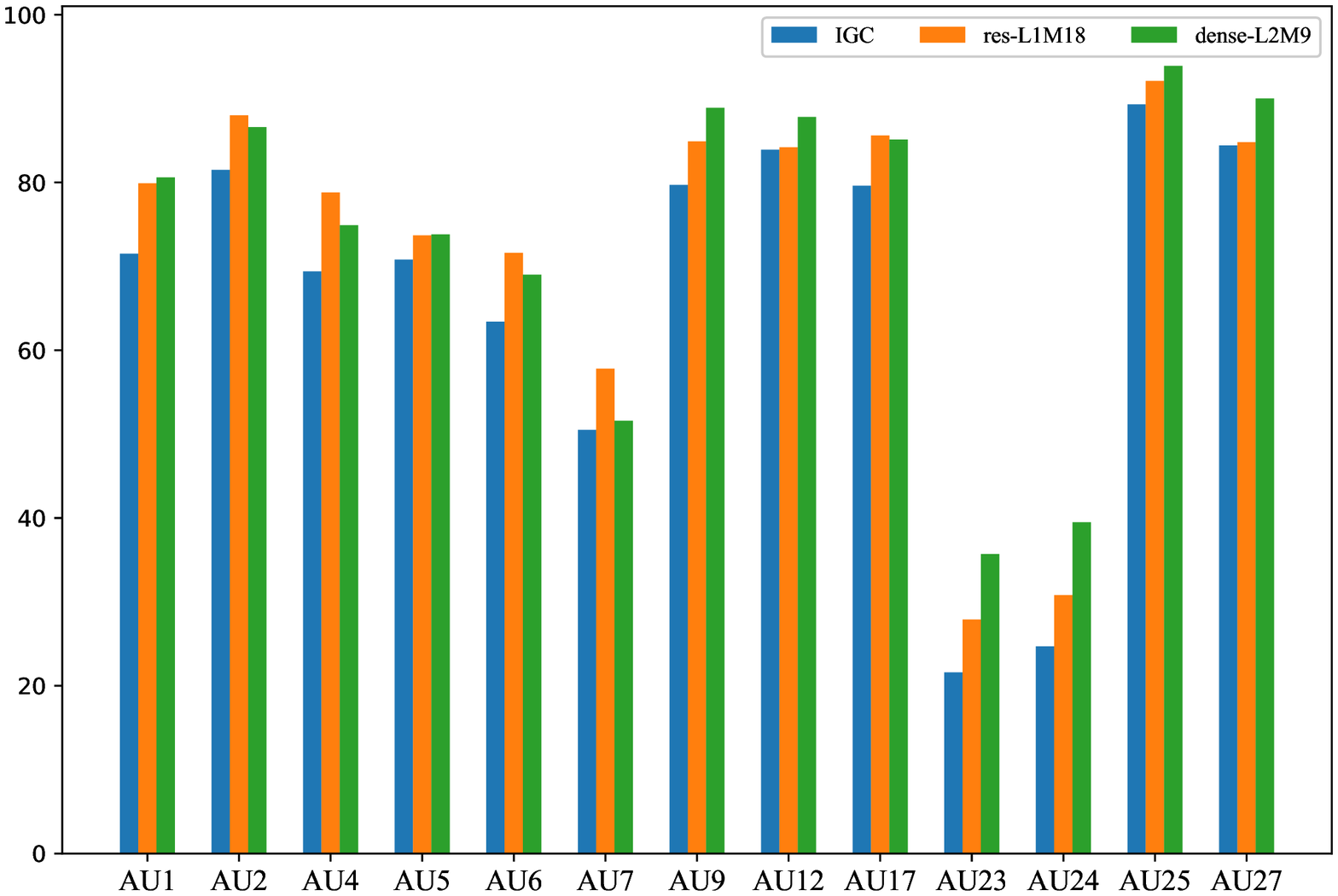}
\caption{The F1 scores of IGC, res-L1M18, dense-L2M9 for each AU on the CK+ dataset}
\label{fig13}
\end{figure}
Table \ref{tab:tab4} and Table \ref{tab:tab5} show the comparison results on DISFA+ and CK+. Compared with the IGC model, the average F1 score has increased by 4.8\% on DISFA+ and increased by 6.7\% on CK+. The results of 6 models (a total of 12 models) of the two connection modes are given in both tables. It can be seen from two tables that the dense connection mode achieves the best results when L=2 and M=9. The average F1 scores of other models are also higher than the corresponding baseline models. This can be attributed to the addition of res connection or dense connection and the subsequent $\text{1}\times \text{1}$ fusion operation after the second group convolution of each module. Res connection and dense connection can not only improve the gradient dissipation problem of the network but also integrate low-level features with high-level features to avoid losing necessary information. The $\text{1}\times \text{1}$ operation solves the problem of increasing the number of feature maps and further realizes the fusion of features. It shows that our proposed network is more suitable for the task of AU detection. For Table \ref{tab:tab4}, although res-L18M1 does not achieve the best results, F1 scores of 5 AUs are the best. Fig.\ref{fig12} shows the F1 scores of IGC, res-L18M1 and dense-L2M9 for each AU on the DISFA+ dataset. In the 12 AUs of the IGC, F1 scores of some AUs have greater than either res-L18M1 or dense-L2M9. But there is no such situation that they are greater than these two models at the same time. And the results are not much different, all within 4.3\%. On AU5, AU6, and AU9, the recognition effect of our proposed network is significantly better than that of the baseline network, with a difference of more than 10\%. Fig.\ref{fig13} shows the F1 scores of IGC, res-L1M18 and dense-L2M9 for each AU on the CK+ dataset. On the CK+ dataset, the model we proposed showed better performance on each AU, especially the results of AU1, AU2, AU23, AU24 and so on. Different models have different recognition capabilities for AU. For example,  res-L1M18 performs better than dense-L2M9 for AU4 in Fig.\ref{fig13}. However, dense-L2M9 performs better than res-L1M18 on AU23 and AU24.  Therefore, without considering the computing power, multiple models can be used to make comprehensive judgments when making inferences.

\begin{table}[htbp]
\centering
\caption{FLOPs and params}
\resizebox{\textwidth}{!}
{
\begin{tabular}{cccccccccccccccc}
\toprule
\multirow{2}{*}{} &\multicolumn{3}{c}{CNN} & \multicolumn{6}{c}{Res connection}&\multicolumn{6}{c}{Dense connection}\\
\cmidrule(r){2-4}\cmidrule(r){5-10}\cmidrule(r){11-16}
& ResNet18&DenseNet121 &IGC &L1M18 & L2M9 & L3M6 & L6M3 & L9M2 & L18M1 & L1M18 & L2M9&L3M6 &L6M3& L9M2&L18M1 \\
\midrule

\tabincell{c}{FLOPs\\($\times {{10}^{9}}$)}& 1.82&2.90& 0.83& 1.13& 0.69& 0.54& 0.40& 0.36& 0.33& 1.32& 0.88& 0.74& 0.60& 0.56& 0.52 \\

\tabincell{c}{Params\\ ($\times $M)}& 11.18& 6.97& 0.57& 13.33& 11.34& 10.68& 10.02& 9.8& 9.59&  14.22& 12.23& 11.57& 10.91& 10.69& 10.48  \\

Depth& 18& 121& 20& \multicolumn{12}{c}{16} \\

\bottomrule

\end{tabular}
}
\label{tab:tab6}
\end{table}
In addition, we also give the floating-point operands per second (FLOPs), the number of parameters and the depth of the baseline networks and CEDNN in Table \ref{tab:tab6}. The results in Table \ref{tab:tab6} are all obtained under the same platform. Table \ref{tab:tab6} shows that as the number of $L$ in our network increases, the number of parameters and FLOPs gradually decreases. When res connection $L=2$ and dense connection $L=3$, FLOPs are smaller than IGC-L24M2. The increased amount of parameters mainly comes from $\text{1}\times \text{1}$ convolution and CNN top. For the depth shown in Table \ref{tab:tab6}, the depth of our network is only 16, and the smallest one of the classic CNN network is 18. Even so, we still get better results than traditional CNN networks. It means that increasing the network width can increase the computing speed and achieve or exceed the effect of a deeper network. Combining the experimental results in Table \ref{tab:tab4} and Table \ref{tab:tab5}, we can conclude that our proposed network can not only ensure that the model complexity is not increased but also achieve a good recognition effect. Compared with traditional CNNs (ResNet, DenseNet, etc.), our network greatly reduces the requirements for experimental hardware, ensuring that experimental results can be obtained quickly on devices with low computing power.

\subsection{Comparison with state-of-the-art method}\label{Comparison with state-of-the-art method}
To validate the effectiveness of our method, we compare our method with the related methods mentioned in Section \ref{Compared Methods}. Due to the availability of the codes for DRML, AU R-CNN and J\^{A}A-Net, we reconduct the experiment of DRML, AU R-CNN and J\^{A}A-Net using our data. The label of the DISFA+ dataset we use is $y\in {{\left\{ 0,1 \right\}}^{d}}$, where $d$ represents the total number of AUs, $0$ indicates that the AU is negative, and $1$ indicates that the AU is positive. The experiment in DRLM article is different from ours, $y\in \{-1,0,1\}$, where $1$ and $-1$ represent positive and negative tags, respectively, and 0 indicates invalid labels. Invalid tags are not included in DISFA+. Therefore, when conducting DRLM comparison experiments, the label was modified to $y\in {{\{-1,1\}}^{d}}$. In the comparative experiment of AU R-CNN, we use the best model, which takes ResNet-101 as the backbone. The input image size is $256\times 256$, the batch size is 24, and the other hyperparameters are the same as those described in the implementation details of \cite{MA201935}. The face data used in the comparison experiment of DISFA+ dataset is obtained by dlib clipping in Section \ref{Spatial Attention Map}. The batch size used in the experiments of DRML and J\^{A}A-Net is 32, and the other hyperparameters are the same as those described in their articles. The model of AU R-CNN is large, and when running on our experimental equipment, the maximum batch size can only be 24. In order to avoid the impact of batch size inconsistencies on the results, we also set the batch size of our model to 24 and get the results of the last column in Table \ref{tab:tab7}. In the comparative experiment of CK+, 12 AUs are used between JPML and DSCMR. We experiment again using the same data, and the results are given in the last column of Table \ref{tab:tab9}.

\begin{table}[htbp]
\centering
\caption{F1 socre result comparison with state-of-the-art methods on DISFA+ dataset. Bracketed bold numbers indicate the best score; bold numbers indicate the second best}
{
\begin{tabular}{cccccc}
\toprule
AU& DRML& AU R-CNN& J\^{A}A-Net& res-L18M1& res-L18M1 (bt24) \\
\midrule
1& 27.3& 47.8& \boldmath $[83.9]$ \unboldmath& \boldmath $83.2$ \unboldmath& 82.3\\

2& 22.2& 42.7& \boldmath $80.5$ \unboldmath& 80.1&\boldmath $[80.7]$ \unboldmath \\

4& 51.0& 55.7& \boldmath $79.3$ \unboldmath& 78.4& \boldmath $[79.7]$ \unboldmath \\

5& 36.4& 47.8& \boldmath $[78.4]$ \unboldmath& 74.3& \boldmath $76.5$ \unboldmath\\

6& 56.2& 42.7& 78.2& \boldmath $[82.3]$ \unboldmath& \boldmath $80.6$ \unboldmath\\

9& 32.0& 24.3& 67.6& \boldmath $[74.7]$ \unboldmath& \boldmath $69.5$ \unboldmath\\

12& 38.5& 47.4& \boldmath $[84.6]$ \unboldmath& \boldmath $83.8$ \unboldmath& 83.6\\

15& 22.7& 23.5& \boldmath $55.4$ \unboldmath& \boldmath $[55.9]$ \unboldmath& 55.1\\

17& 27.1& 6.4& 60.4& \boldmath $[65.0]$ \unboldmath& \boldmath $64.8$ \unboldmath\\

20& 16.3& 28.7& 48.5& \boldmath $49.7$ \unboldmath& \boldmath $[51.8]$ \unboldmath\\

25& 56.5& 53.1& 85.1& \boldmath $88.2$ \unboldmath& \boldmath $[89.9]$ \unboldmath\\

26& 42.3& 38.6& 69.0& \boldmath $[76.6]$ \unboldmath& \boldmath $74.1$ \unboldmath\\

Avg& 35.7& 38.2& 72.6& \boldmath $[74.4]$ \unboldmath& \boldmath $74.0$ \unboldmath\\

\bottomrule
\end{tabular}
}
\label{tab:tab7}
\end{table}

The experimental results on the DISFA+ dataset are shown in Table \ref{tab:tab7}. It can be seen from Table \ref{tab:tab7} that our CEDNN is superior to all the state-of-the-art works. Specifically, the average F1 score of res-L18M1 is 1.8\% higher than that of J\^{A}A-Net. Because of the serious data imbalance in DISFA+, the performance of different AUs will oscillate seriously in most previous methods. The J\^{A}A-Net weighs the loss of each AU. On the contrary, our method does not use loss weighting and achieves better results. The F1 scores of AU1, AU5 and AU12 of J\^{A}A-Net are higher than those of res-L18M1. But the difference between the results is small, with a maximum difference of 4.1\% for AU5. On the other hand, the F1 score difference between res-L18M1 and J\^{A}A-Net for AU9 (AU26) can reach more than 7\%. The result of the AU R-CNN is not satisfactory. In the comparative experiment of basic convolution network, the result of using ResNet-101 is worse than that of ResNet18. With the deepening of the network, the parameters of the model will be greatly increased. After reaching a certain number of layers, it tends to be saturated, so the generalization ability begins to decline. In other words, performance is not necessarily proportional to network depth. That can also explain why AU R-CNN is less effective.
\begin{table}[htbp]
\centering
\caption{Inference time (ms), parameters and FLOPs of comparison methods on DISFA+ dataset.}
{
\begin{tabular}{cccc}
\toprule
&  AU R-CNN& J\^{A}A-Net& res-L18M1 \\
\midrule
\tabincell{c}{Inference\\ time}& $27.1\pm 3.52\text{E}-06$& $72.8\pm 1.86\text{E}-06$& $8.08\pm 8.29\text{E}-07$\\

\tabincell{c}{FLOPs\\($\times {{10}^{9}}$)}& -&281.8& 0.33\\

\tabincell{c}{Params\\ ($\times $M)}& -& 25.24& 9.79  \\

\bottomrule
\end{tabular}
}
\label{tab:tab8}
\end{table}

We further evaluated inference time, parameters, and FLOPs among CEDNN, J\^{A}A-Net, and AU R-CNN. We tracked each network 12 times. When the same image is input, the average value and standard deviation of 12 trials are calculated. The specific data is shown in Table \ref{tab:tab8}. The inference time of our method is significantly lower than that of AU R-CNN and J\^{A}A-Net. In particular, the inference time of J\^{A}A-Net is seven times longer than that of our method. In our method, each Block is composed of group convolution. Compared with the conventional convolution operation, the number of parameters and computation cost of group convolution is relatively low. J\^{A}A-Net is composed of 6 network modules and uses conventional convolution, which leads to a huge amount of network parameters and an increase in computational cost. These six modules use series-parallel mode, and frequent interaction increases inference time.

\begin{table}[htp]
\centering
\caption{F1 socre result comparison with state-of-the-art methods on CK+ dataset. Bracketed bold numbers indicate the best score; bold numbers indicate the second best}
\resizebox{\textwidth}{!}
{
\begin{tabular}{cccccccccc}
\toprule
AU&BGCS&HRBM&\textbf{res-L3M6}&JPML&DSCMR&L-DSCMR&\textbf{\tabincell{c}{res-L3M6\\(12 classes)}}\\
\midrule
1&71.0&\boldmath $86.9$ \unboldmath&85.9&50.0&54.2&64.3&\boldmath $[87.5]$ \unboldmath\\
2&66.9&85.5&\boldmath $[88.4]$ \unboldmath&40.0&64.7&58.6&\boldmath $86.1$ \unboldmath\\
4&63.6&72.6&\boldmath $80.7$ \unboldmath&72.3&61.1&69.1&\boldmath $[82.1]$ \unboldmath\\
5&59.6&72.0&\boldmath $74.2$ \unboldmath&52.6&42.2&66.7&\boldmath $[74.9]$ \unboldmath\\
6&57.8&61.7&\boldmath $[70.0]$ \unboldmath&57.8&\boldmath $68.2$ \unboldmath&62.2&68.1\\
7&46.6&54.5&\boldmath $[61.6]$ \unboldmath&24.4&35.9&47.1&\boldmath $55.8$ \unboldmath\\
9&51.9&85.9&\boldmath $89.2$ \unboldmath&54.5&53.8&51.5&\boldmath $[89.9]$ \unboldmath\\
12&73.4&72.7&\boldmath $87.2$ \unboldmath&75.6&80.0&75.6&\boldmath $[87.3]$ \unboldmath\\
17&76.5&81.7&\boldmath $86.4$ \unboldmath&81.9&\boldmath $[89.7]$ \unboldmath&82.4&84.6\\
23&22.2&56.6&45.8&41.7&\boldmath $[75.0]$ \unboldmath&\boldmath $73.3$ \unboldmath&32.6\\
24&23.9&35.3&\boldmath $[46.2]$ \unboldmath&30.8&36.4&26.7&\boldmath $43.2$ \unboldmath\\
25&78.5&\boldmath $92.6$ \unboldmath&\boldmath $[93.6]$ \unboldmath&76.4&85.5&89.1&91.4\\
27&57.9&\boldmath $87.7$ \unboldmath&\boldmath $[89.6]$ \unboldmath&-&-&-&-\\
Avg&57.7&72.7&\boldmath $[76.8]$ \unboldmath&54.8&62.2&63.9&\boldmath $73.6$ \unboldmath \\

\bottomrule
\end{tabular}
}
\label{tab:tab9}
\end{table}

Table \ref{tab:tab9} reports F1 scores for different methods on CK+. As you can see, our approach is superior to all previous work on CK+ datasets. When using 13 AUs, compared with the traditional BGCS and HRBM methods, the average F1 score of CEDNN increased by 19.1\% and 4.1\%, respectively. It shows the superiority of the method based on deep learning. When using 12 AUs, the average F1 score of res-L3M6 increased by 18.8\% and 11.4\%, respectively, compared with deep learning methods such as JPML and DSCMR. Except for AU17 and AU23, the F1 score of our method is higher or much higher than other methods.

\subsection{Ablation study}\label{Ablation study}
\begin{figure}[h]
  \centering
  \subfigure []{
  \includegraphics[width=4cm]{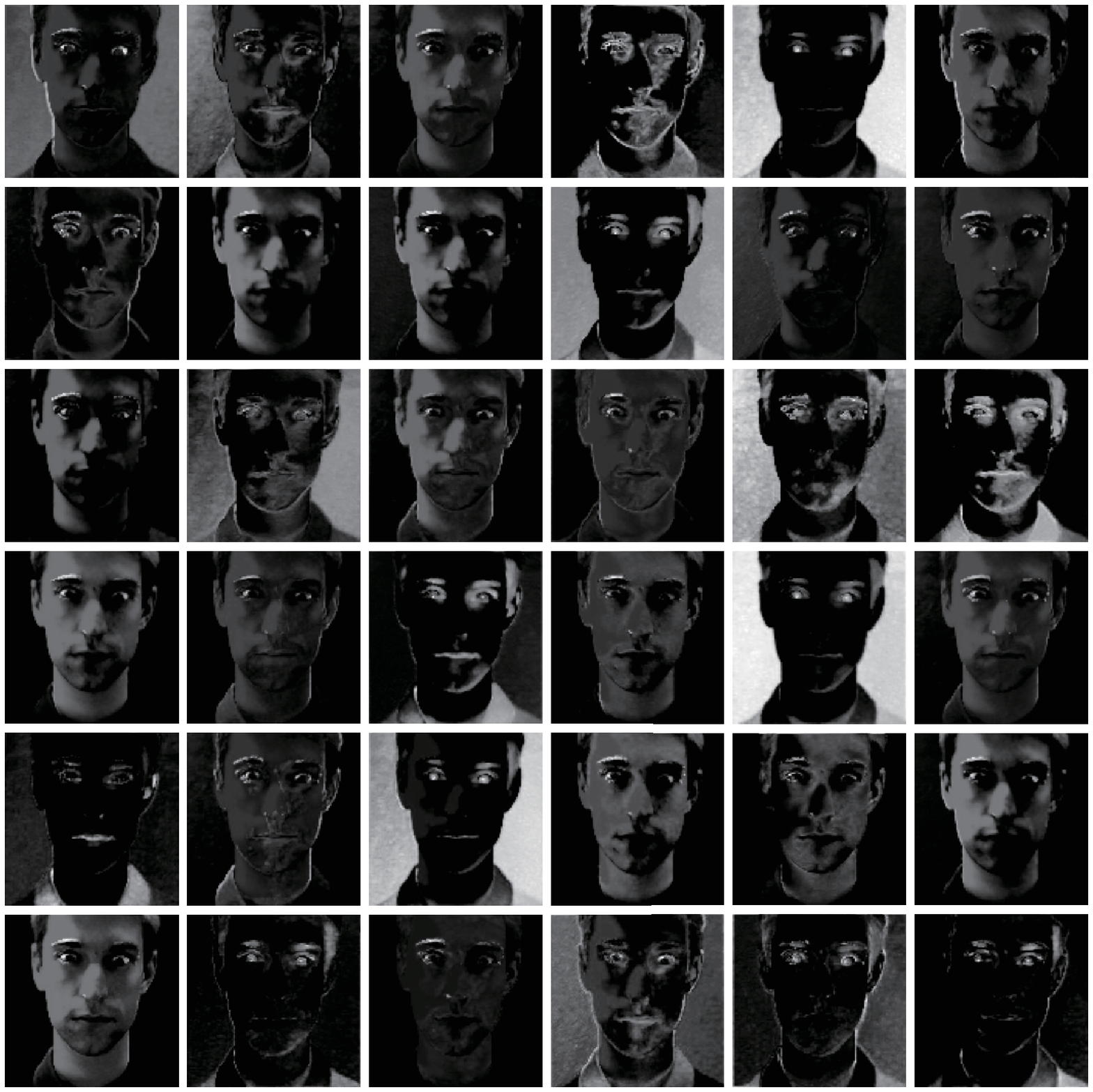}
  }
  \subfigure []{
  \includegraphics[width=4cm]{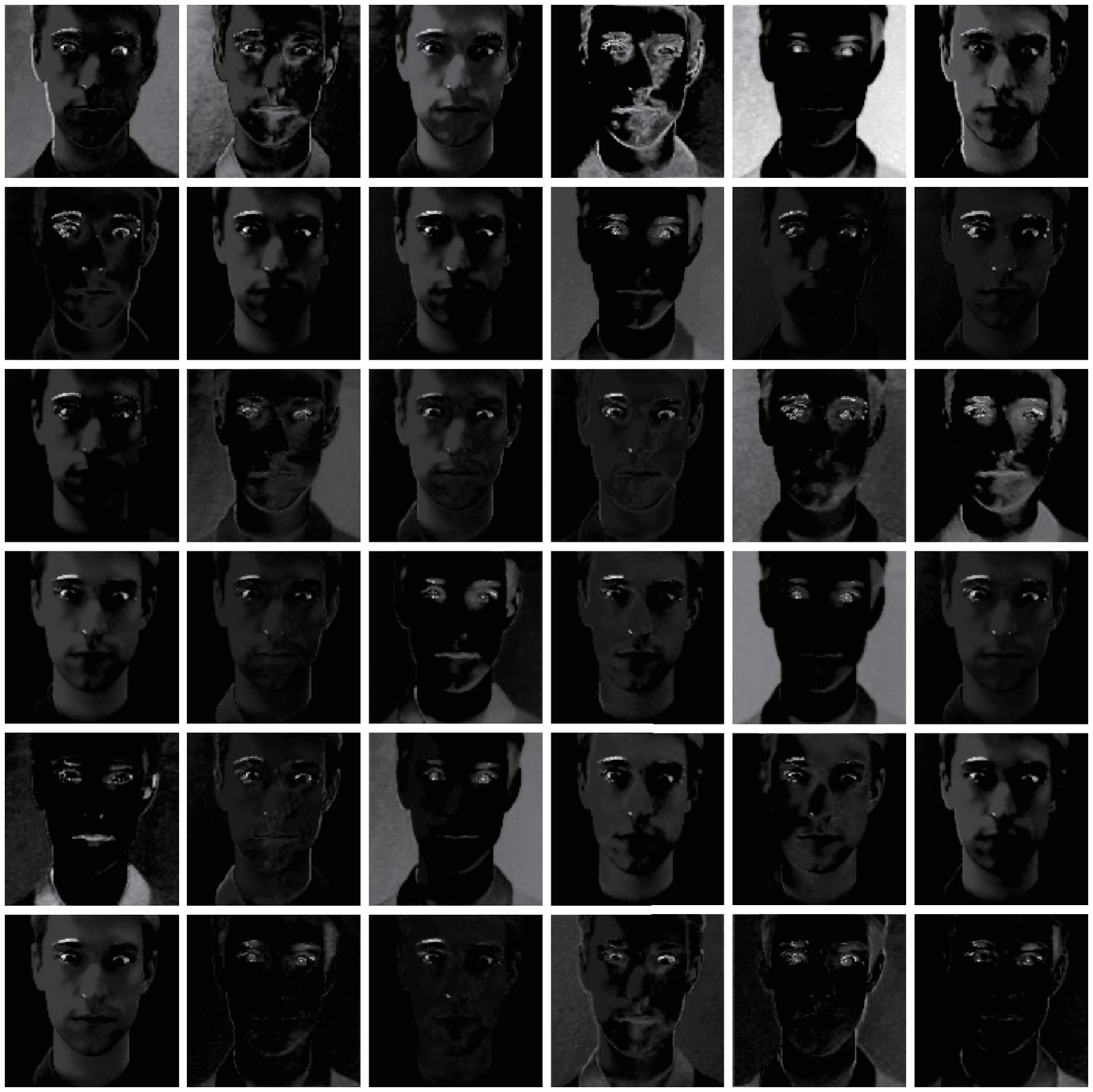}
  }
  \caption{Visualization results of Block1 feature map before and after adding spatial attention}
  \label{fig14}
\end{figure}

From the analysis in Section \ref{conventional CNN versus CEDNN}, it can be seen that CEDNN still achieves better results than the baseline network when the spatial attention map is not added. The spatial attention map we proposed in Section \ref{Spatial Attention Map} aims to enhance the pixels in certain regions to achieve the purpose of identifying local regions AU. Tables \ref{tab:tab10} and \ref{tab:tab15} show the F1 score of each model after adding the spatial attention map. By comparing the data in Table \ref{tab:tab10} and Table \ref{tab:tab4} (Table \ref{tab:tab15} and Table \ref{tab:tab5}), the addition of spatial attention map does improve the overall recognition rate. In particular, the mean F1 score of the res-L3M6 model has an overall improvement of 5.1\% on the DISFA+ dataset. The average F1 score of the res-L3M6 model has an improvement of 3.5\% on the CK+ dataset. We visualize the feature map extracted by the first Block, as shown in Fig.\ref{fig14}(a). The visualization results after adding the spatial attention map are shown in Fig.\ref{fig14}(b). From the second row to the sixth row, we can see that the attention map we proposed can enhance the focus on certain regions at the feature map level. As described in Section \ref{channel attention}, adding channel attention can enhance channels related to key information. The results of two se connection methods on the DISFA+ dataset are shown in Table \ref{tab:tab11} and Table \ref{fig12}. Through comparison, it can be found that the first connection method of se connection generally increases the average F1 score by 1\%-2\%. Compared with the first type of connection, the F1 score of each model of the second type of se connection is not significantly improved. That may be because the extracted features will be different after passing through the se-module. However, after adding or connecting with the input of the Block, the influence of the se-module will be weakened or even canceled. So the improvement effect is not obvious. The feature map extracted by the entire block can be sent to the se-module to enhance the channel containing more effective information more effectively.

\begin{table}[htbp]
\centering
\caption{Control experiments for DISFA+. Results of model with attention maps are reported using F1 score on 3-fold protocol.}
\resizebox{\textwidth}{!}
{
\begin{tabular}{ccccccccccccc}
\toprule
\multirow{2}{*}{AU} & \multicolumn{6}{c}{Res connection}&\multicolumn{6}{c}{Dense connection}\\
\cmidrule(r){2-7}\cmidrule(r){8-13}
 &L1M18 & L2M9 & L3M6 & L6M3 & L9M2 & L18M1 & L1M18 & L2M9&L3M6 &L6M3& L9M2&L18M1 \\
\midrule

1&83.1&81.9&81.4&82.2&80.9&\boldmath $84.5$ \unboldmath&81.4&82.9&81.5&81.5&81.9&82.6\\
2&79.8&80.7&80.7&79.8&79.2&\boldmath $82.4$ \unboldmath&80.2&81.1&80.6&79.9&81.0&82.2\\
4&77.9&79.1&78.1&82.0&79.7&\boldmath $83.9$ \unboldmath&80.1&81.4&78.5&78.5&79.8&81.4\\
5&76.5&72.5&75.1&73.8&75.9&72.8&76.2&74.8&75.8&74.0&73.8&\boldmath $76.9$ \unboldmath\\
6&\boldmath $85.4$ \unboldmath&80.8&84.6&85.0&84.2&84.4&84.5&83.5&84.6&84.2&83.6&83.9\\
9&67.4&68.9&71.6&69.2&67.5&67.2&71.0&70.1&70.3&\boldmath $72.4$ \unboldmath&68.0&69.0\\
12&82.8&77.8&84.7&83.1&81.5&\boldmath $84.8$ \unboldmath&83.5&84.5&82.6&83.6&81.5&82.1\\
15&48.7&51.0&\boldmath $53.3$ \unboldmath&49.9&\boldmath $53.3$ \unboldmath&46.4&42.1&50.6&49.6&50.5&47.2&37.8\\
17&56.9&57.3&\boldmath $67.0$ \unboldmath&61.8&62.9&61.1&59.7&59.3&52.9&60.9&61.1&60.5\\
20&39.0&50.8&48.3&47.1&45.9&\boldmath $55.0$ \unboldmath&42.3&43.2&38.8&39.1&51.5&51.0\\
25&86.7&88.0&87.1&87.6&87.6&87.6&85.0&83.9&87.1&\boldmath $89.6$ \unboldmath&87.2&87.8\\
26&74.7&\boldmath $78.1$ \unboldmath&75.8&75.5&72.9&75.2&75.2&73.0&75.6&76.1&75.7&77.9\\
Avg&71.6&72.2&\boldmath $74.0$ \unboldmath&73.1&72.6&73.8&71.8&72.4&71.5&72.5&72.7&72.8\\
\bottomrule

\end{tabular}
}
\label{tab:tab10}
\end{table}

\begin{table}[htbp]
\centering
\caption{Control experiments for DISFA+. Results of model with attention maps and first se-block are reported using F1 score on 3-fold protocol.}
\resizebox{\textwidth}{!}
{
\begin{tabular}{ccccccccccccc}
\toprule
\multirow{2}{*}{AU} & \multicolumn{6}{c}{Res connection}&\multicolumn{6}{c}{Dense connection}\\
\cmidrule(r){2-7}\cmidrule(r){8-13}
 &L1M18 & L2M9 & L3M6 & L6M3 & L9M2 & L18M1 & L1M18 & L2M9&L3M6 &L6M3& L9M2&L18M1 \\
\midrule

1&82.9&80.3&82.0&\boldmath $84.6$ \unboldmath&82.3&82.0&81.2&82.2&82.9&80.5&80.4&81.9\\
2&78.9&79.2&81.4&78.4&80.2&78.5&78.8&81.1&\boldmath $81.8$ \unboldmath&81.4&79.3&79.3\\
4&77.9&81.7&79.7&78.5&77.7&81.2&77.7&\boldmath $83.3$ \unboldmath&80.0&79.7&78.2&79.0\\
5&77.4&75.2&77.3&74.2&74.8&76.2&77.4&75.2&\boldmath $77.6$ \unboldmath&76.6&74.2&74.8\\
6&\boldmath $85.2$ \unboldmath&84.0&82.6&84.6&83.8&80.9&83.0&83.1&83.1&\boldmath $85.2$ \unboldmath&84.6&84.7\\
9&70.3&69.7&72.2&68.8&66.3&69.7&66.3&71.4&\boldmath $73.5$ \unboldmath&69.9&72.7&67.6\\
12&83.8&\boldmath $84.6$ \unboldmath&81.6&82.3&82.8&82.3&83.8&83.2&83.8&82.9&81.9&83.5\\
15&51.0&49.7&52.8&47.5&49.9&51.7&50.1&\boldmath $53.2$ \unboldmath&53.0&49.5&52.8&44.4\\
17&60.8&64.2&61.3&58.3&62.7&58.6&62.8&64.5&57.4&58.6&\boldmath $65.0$ \unboldmath&\boldmath $65.0$ \unboldmath\\
20&41.5&48.3&46.0&\boldmath $53.0$ \unboldmath&43.2&48.3&47.5&42.9&43.3&47.9&47.6&42.9\\
25&88.8&86.4&87.3&88.1&88.4&87.5&\boldmath $88.9$ \unboldmath&87.7&87.3&88.4&85.6&87.0\\
26&75.3&74.3&74.2&75.1&76.1&72.1&\boldmath $78.1$ \unboldmath&74.4&74.2&76.7&74.4&77.6\\
Avg&72.8&73.1&73.2&72.8&72.4&72.4&73.0&\boldmath $73.5$ \unboldmath&73.2&73.1&73.1&72.3\\
\bottomrule

\end{tabular}
}
\label{tab:tab11}
\end{table}

\begin{table}[htbp]
\centering
\caption{Control experiments for DISFA+. Results of model with attention maps and second se-block are reported using F1 score on 3-fold protocol.}
\resizebox{\textwidth}{!}
{
\begin{tabular}{ccccccccccccc}
\toprule
\multirow{2}{*}{AU} & \multicolumn{6}{c}{Res connection}&\multicolumn{6}{c}{Dense connection}\\
\cmidrule(r){2-7}\cmidrule(r){8-13}
 &L1M18 & L2M9 & L3M6 & L6M3 & L9M2 & L18M1 & L1M18 & L2M9&L3M6 &L6M3& L9M2&L18M1 \\
\midrule
1&81.6&81.0&82.0&\boldmath $82.9$ \unboldmath&82.5&82.4&81.6&\boldmath $82.5$ \unboldmath&81.5&82.3&81.8&80.5\\
2&79.9&79.8&80.7&\boldmath $81.7$ \unboldmath&81.2&80.1&81.3&81.3&80.1&78.4&79.4&79.1\\
4&78.8&79.3&\boldmath $81.2$ \unboldmath&77.2&77.4&81.0&78.6&79.5&77.4&79.3&79.7&80.5\\
5&71.7&76.0&76.6&73.6&76.7&75.2&70.6&76.5&72.8&72.3&\boldmath $77.5$ \unboldmath&72.6\\
6&79.5&84.9&84.8&84.9&83.3&83.3&83.5&83.4&\boldmath $85.0$ \unboldmath&83.7&82.2&83.9\\
9&69.0&67.2&66.4&63.3&65.7&71.1&71.4&\boldmath $71.7$ \unboldmath&61.4&63.6&66.6&68.4\\
12&82.5&81.3&82.3&81.8&83.2&82.0&\boldmath $83.4$ \unboldmath&82.4&82.1&81.1&82.6&81.8\\
15&44.8&45.4&50.1&47.2&\boldmath $54.3$ \unboldmath&40.6&40.1&44.5&47.2&40.6&43.5&53.2\\
17&54.6&59.8&60.3&60.8&64.5&60.6&\boldmath $65.5$ \unboldmath&56.2&63.2&58.1&60.8&59.9\\
20&43.4&46.6&45.9&47.0&39.9&42.7&44.3&\boldmath $52.2$ \unboldmath&51.9&42.1&38.2&40.3\\
25&86.5&86.1&87.6&86.1&87.3&85.2&86.3&85.0&\boldmath $88.3$ \unboldmath&86.8&86.4&85.7\\
26&73.3&\boldmath $79.0$ \unboldmath&74.3&77.2&77.7&73.6&74.3&73.8&76.1&73.2&75.2&75.8\\
avg&70.5&72.2&72.7&72.0&\boldmath $72.8$ \unboldmath&71.5&71.7&72.4&72.3&70.1&71.2&71.8\\
\bottomrule
\end{tabular}
}
\label{tab:tab12}
\end{table}

\begin{table}[htbp]
\centering
\caption{Control experiments for DISFA+. Results of model with three-layers attention maps and first se-block are reported using F1 score on 3-fold protocol.}
\resizebox{\textwidth}{!}
{
\begin{tabular}{ccccccccccccc}
\toprule
\multirow{2}{*}{AU} & \multicolumn{6}{c}{Res connection}&\multicolumn{6}{c}{Dense connection}\\
\cmidrule(r){2-7}\cmidrule(r){8-13}
 &L1M18 & L2M9 & L3M6 & L6M3 & L9M2 & L18M1 & L1M18 & L2M9&L3M6 &L6M3& L9M2&L18M1 \\
\midrule
1&82.9&83.6&81.6&81.5&\boldmath $83.9$ \unboldmath&83.2&78.3&83.0&82.9&82.4&82.6&82.7\\
2&\boldmath $82.7$ \unboldmath&80.1&79.1&81.0&81.5&80.1&78.3&79.2&79.3&81.1&80.5&80.1\\
4&79.9&81.9&81.2&78.3&79.4&78.4&81.6&79.6&81.7&79.4&\boldmath $82.1$ \unboldmath&\boldmath $82.1$ \unboldmath\\
5&\boldmath $78.2$ \unboldmath&72.6&75.3&75.7&75.9&74.3&75.9&74.6&73.3&77.4&75.7&75.4\\
6&80.7&\boldmath $85.9$ \unboldmath&83.7&82.4&80.2&82.3&84.5&82.7&80.7&83.9&83.2&85.8\\
9&71.6&66.2&73.3&73.0&69.5&\boldmath $74.7$ \unboldmath&72.3&69.2&74.0&67.1&74.2&65.7\\
12&79.9&\boldmath $84.9$ \unboldmath&83.8&83.9&83.6&83.8&82.4&83.7&81.3&82.5&80.7&83.4\\
15&55.8&50.1&54.6&54.2&\boldmath $57.2$ \unboldmath&55.9&53.3&52.8&54.5&49.2&50.7&50.1\\
17&64.8&65.7&64.1&64.9&58.5&65.0&64.6&\boldmath $66.5$ \unboldmath&62.4&58.8&56.4&63.6\\
20&51.9&43.4&43.1&48.5&40.5&49.7&\boldmath $54.5$ \unboldmath&45.8&53.1&42.0&51.0&48.2\\
25&86.3&\boldmath $89.2$ \unboldmath&85.5&87.8&87.2&88.2&87.2&87.4&87.0&87.5&87.8&84.8\\
26&74.6&74.5&69.7&75.1&72.7&\boldmath $76.6$ \unboldmath&71.0&75.9&72.2&73.6&73.9&72.7\\
Avg&74.1&73.2&72.9&73.9&72.5&\boldmath $74.4$ \unboldmath&73.6&73.4&73.5&72.1&73.2&72.9\\
\bottomrule
\end{tabular}
}
\label{tab:tab13}
\end{table}

\begin{table}[htbp]
\centering
\caption{Control experiments for DISFA+. Results of model with four-layers attention maps and first se-block are reported using F1 score on 3-fold protocol}
\resizebox{\textwidth}{!}
{
\begin{tabular}{ccccccccccccc}
\toprule
\multirow{2}{*}{AU} & \multicolumn{6}{c}{Res connection}&\multicolumn{6}{c}{Dense connection}\\
\cmidrule(r){2-7}\cmidrule(r){8-13}
 &L1M18 & L2M9 & L3M6 & L6M3 & L9M2 & L18M1 & L1M18 & L2M9&L3M6 &L6M3& L9M2&L18M1 \\
\midrule
1&80.4&80.3&80.5&82.5&79.7&80.9&\boldmath $84.3$ \unboldmath&81.7&81.0&81.1&82.1&81.8\\
2&79.1&79.5&79.6&78.4&77.3&78.3&78.6&78.2&79.6&78.2&\boldmath $80.9$ \unboldmath&76.2\\
4&78.2&79.3&79.5&76.0&78.2&78.1&77.1&76.0&76.4&\boldmath $79.8$ \unboldmath&79.2&78.1\\
5&75.6&76.3&73.9&74.8&75.9&76.1&76.0&73.4&\boldmath $77.8$ \unboldmath&74.1&74.1&75.4\\
6&81.9&84.1&81.9&84.4&\boldmath $85.4$ \unboldmath&82.7&83.9&82.8&81.8&84.5&84.0&84.8\\
9&\boldmath$74.4$ \unboldmath&73.2&68.7&66.9&70.9&73.2&70.9&68.6&68.2&67.8&73.6&63.5\\
12&84.2&84.6&83.0&82.0&82.0&\boldmath $84.8$ \unboldmath&84.2&83.8&83.0&83.5&84.2&82.3\\
15&51.7&51.4&51.6&51.6&40.5&51.0&49.0&\boldmath $53.4$ \unboldmath&52.0&51.6&46.5&52.1\\
17&55.1&59.5&63.7&60.9&63.3&64.7&64.2&59.6&57.3&62.8&58.3&\boldmath $66.4$ \unboldmath\\
20&42.3&36.7&35.7&47.6&49.0&39.7&45.0&\boldmath $52.8$ \unboldmath&49.4&50.5&44.9&43.7\\
25&87.2&86.2&86.1&86.7&88.0&86.3&87.7&85.4&87.0&84.7&88.2&\boldmath $89.0$ \unboldmath\\
26&74.5&76.8&75.9&\boldmath $78.3$ \unboldmath&76.3&72.8&75.8&74.9&76.7&74.6&75.1&74.2\\
Avg&72.0&72.3&71.7&72.5&72.2&72.4&\boldmath $73.1$ \unboldmath&72.6&72.5&72.8&72.6&72.3\\
\bottomrule
\end{tabular}
}
\label{tab:tab14}
\end{table}

\begin{table}[htbp]
\centering
\caption{Control experiments for CK+. Results of model with three-layers attention maps are reported using F1 score}
\resizebox{\textwidth}{!}
{
\begin{tabular}{ccccccccccccc}
\toprule
\multirow{2}{*}{AU} & \multicolumn{6}{c}{Res connection}&\multicolumn{6}{c}{Dense connection}\\
\cmidrule(r){2-7}\cmidrule(r){8-13}
 &L1M18 & L2M9 & L3M6 & L6M3 & L9M2 & L18M1 & L1M18 & L2M9&L3M6 &L6M3& L9M2&L18M1 \\
\midrule
1&85.3&86.4&85.9&84.4&87.0&83.1&86.8&83.1&86.1&84.3&\boldmath $88.2$ \unboldmath&84.7\\
2&88.2&88.1&88.4&\boldmath $89.7$ \unboldmath&88.8&87.1&86.9&87.6&87.8&86.0&86.5&87.3\\
4&80.4&81.0&80.7&81.6&80.8&80.7&80.3&\boldmath $82.3$ \unboldmath&79.5&78.4&80.8&77.4\\
5&74.5&73.6&74.2&75.9&77.2&74.4&77.4&\boldmath $79.8$ \unboldmath&76.3&76.8&72.4&72.8\\
6&72.4&69.6&70.0&\boldmath $73.9$ \unboldmath&68.7&71.2&70.9&73.0&71.5&69.6&68.9&\boldmath $73.9$ \unboldmath\\
7&51.8&57.4&61.6&59.3&57.1&54.8&\boldmath $61.7$ \unboldmath&57.8&54.9&55.6&58.4&58.2\\
9&87.3&91.8&89.2&89.2&90.5&90.3&87.8&90.4&\boldmath $93.2$ \unboldmath&90.9&88.1&88.3\\
12&\boldmath $88.8$ \unboldmath&87.9&87.2&86.7&85.4&86.3&84.9&85.8&86.5&86.0&86.5&88.3\\
17&85.6&86.7&86.4&84.1&86.4&84.7&\boldmath $87.0$ \unboldmath&84.5&85.6&83.7&85.9&84.5\\
23&38.3&37.0&\boldmath $45.8$ \unboldmath&40.9&40.4&35.0&40.4&41.3&38.2&32.3&29.2&34.1\\
24&40.4&46.7&46.2&42.4&\boldmath $51.1$ \unboldmath&42.7&43.4&40.4&38.6&43.2&42.4&35.3\\
25&92.8&92.7&93.6&92.7&91.8&91.7&91.5&91.5&93.9&92.5&91.7&\boldmath $94.2$ \unboldmath\\
27&\boldmath $92.7$ \unboldmath&92.0&89.6&90.4&92.0&89.7&91.4&91.5&90.8&90.2&90.8&91.5\\
Avg&75.3&76.2&\boldmath $76.8$ \unboldmath&76.2&76.7&74.7&76.2&76.1&75.6&74.6&74.6&74.6\\
\bottomrule
\end{tabular}
}
\label{tab:tab15}
\end{table}

\begin{table}[htbp]
\centering
\caption{Control experiments for CK+. Results of model with three-layers attention maps and first se-block are reported using F1 score.}
\resizebox{\textwidth}{!}
{
\begin{tabular}{ccccccccccccc}
\toprule
\multirow{2}{*}{AU} & \multicolumn{6}{c}{Res connection}&\multicolumn{6}{c}{Dense connection}\\
\cmidrule(r){2-7}\cmidrule(r){8-13}
 &L1M18 & L2M9 & L3M6 & L6M3 & L9M2 & L18M1 & L1M18 & L2M9&L3M6 &L6M3& L9M2&L18M1 \\
\midrule
1&\boldmath $88.5$ \unboldmath&86.0&85.0&86.0&85.6&85.8&88.0&85.0&87.1&85.0&86.1&84.7\\
2&87.8&86.7&87.9&86.6&88.0&\boldmath $88.4$ \unboldmath&86.2&86.9&86.6&86.3&86.9&85.7\\
4&81.3&81.4&81.7&80.5&80.5&81.8&79.9&80.8&79.7&81.1&80.8&\boldmath $82.0$ \unboldmath\\
5&70.9&75.6&76.7&71.3&77.6&76.2&75.4&73.3&74.7&\boldmath $78.4$ \unboldmath&74.2&74.3\\
6&71.9&68.4&68.7&\boldmath $72.4$ \unboldmath&70.1&70.6&70.9&72.2&71.9&67.8&65.8&72.2\\
7&59.7&60.3&58.7&59.1&55.0&\boldmath $61.0$ \unboldmath&58.6&51.1&54.7&56.8&58.0&54.3\\
9&88.3&89.4&91.0&88.3&90.2&88.1&88.4&89.0&\boldmath $91.7$ \unboldmath&89.4&87.3&89.2\\
12&87.9&88.3&86.7&88.2&88.3&87.6&88.2&84.0&89.1&87.8&\boldmath $90.6$ \unboldmath&87.4\\
17&85.3&84.6&86.7&86.4&84.9&82.9&84.7&\boldmath $87.7$ \unboldmath&85.9&85.6&86.0&85.0\\
23&41.8&41.7&38.8&40.8&34.4&32.7&31.8&31.1&29.2&39.6&28.9&\boldmath $49.5$ \unboldmath\\
24&40.9&37.6&\boldmath $50.0$ \unboldmath&41.9&37.0&40.9&42.0&39.5&40.0&39.5&35.0&31.7\\
25&92.2&91.5&91.4&91.8&\boldmath $93.4$ \unboldmath&92.2&91.8&92.1&92.6&92.3&92.4&91.2\\
27&90.0&90.7&88.8&90.6&90.8&91.9&88.2&91.3&90.0&\boldmath $92.0$ \unboldmath&89.4&88.2\\
Avg&75.9&75.6&\boldmath $76.3$ \unboldmath&75.7&75.1&75.4&74.9&74.2&74.9&75.5&74.0&75.0\\
\bottomrule
\end{tabular}
}
\label{tab:tab16}
\end{table}

\begin{figure}[htp]
\centering
\includegraphics[width=13cm]{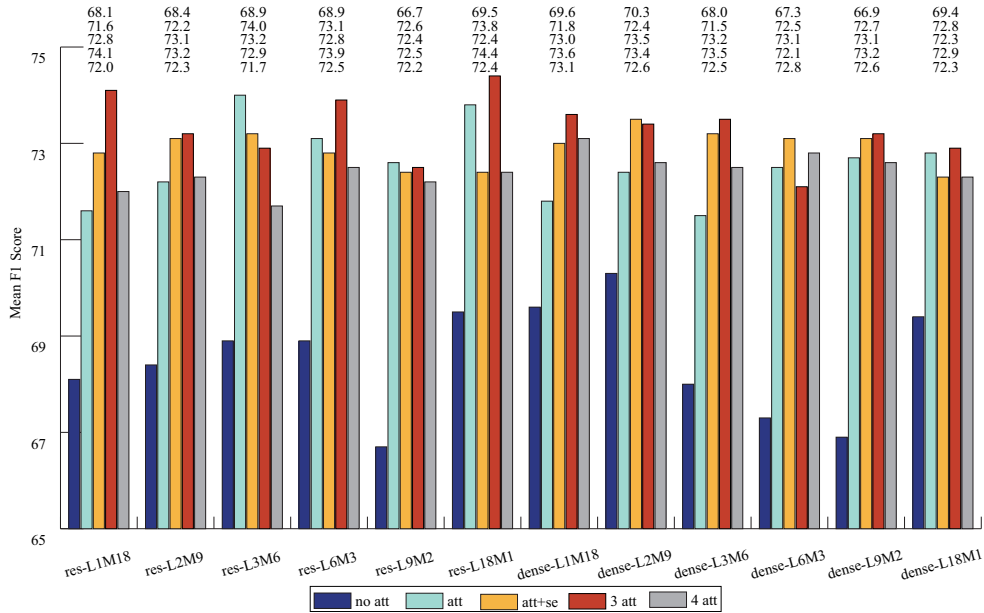}
\caption{Average F1 score of each model on DISFA+}
\label{fig15}
\end{figure}

After verifying the effectiveness of adding the channel attention module, we determine the number of layers to add to the spatial attention map. The last column in the Table \ref{tab:tab3} shows whether each block adds spatial attention maps. In our model, $L\times M$ is always a multiple of 6, so spatial attention maps can be added to the input of each block. Spatial attention maps are added to the first two blocks in Tables \ref{tab:tab10}, \ref{tab:tab11} and \ref{tab:tab12}. Tables \ref{tab:tab13} and \ref{tab:tab14} show the results of adding three-layer and four-layer attention maps to each model.  Since there are many model results displayed, for a more intuitive comparison, we show the average F1 score results of each model in Fig.\ref{fig15}. In the legend of Fig.\ref{fig15}, no-att shows the result when there is no spatial attention map in Table \ref{tab:tab4}. The result of adding the spatial attention map is represented by att. The result of adding the first se connection mode is represented by att+se. 3 att shows the result of adding three-layer spatial attention maps and using the first se-block. 4 att shows the result of adding four-layer spatial attention maps and using the first se-block.

It can be seen directly from Fig.\ref{fig15} that the models without spatial attention are much worse than the results of their corresponding models. Most models show a trend of increasing scores. When the spatial attention map is added to four layers, the result is worse than adding three layers. It can be seen that more isn't always better. As the depth increases, the extracted features become more and more abstract. The spatial attention map can play a role in highlighting key areas on the image layer, and loses its effect when feature map reaches a certain semantic feature.

\begin{figure}[htp]
\centering
\includegraphics[width=13cm]{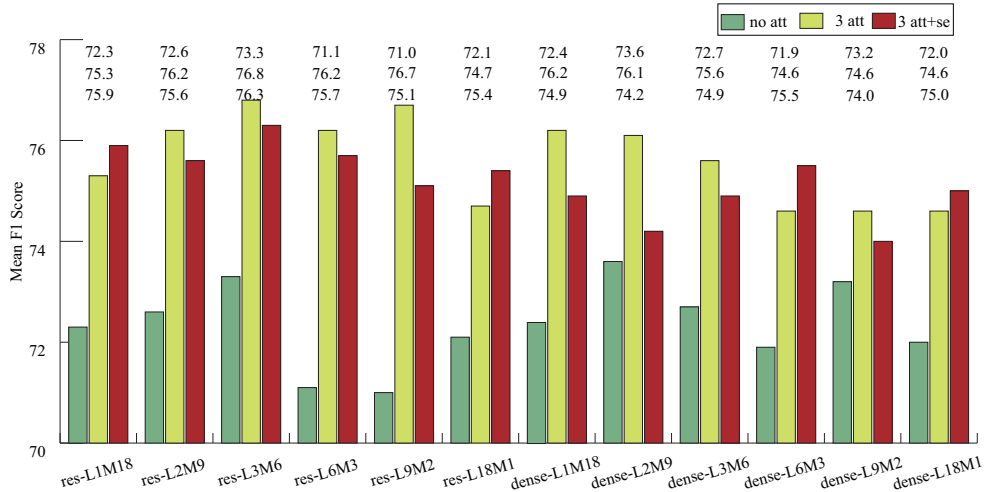}
\caption{Average F1 score of each model on CK+}
\label{fig16}
\end{figure}

On the CK+ dataset, the results of adding three layers of attention maps are shown in Table \ref{tab:tab15}, The results of adding the three-layer attention map and the first se connection are shown in Table \ref{tab:tab16}. The results of the average F1 scores of each model on the CK+ dataset are shown in Fig.\ref{fig16}. It can be seen from Fig.\ref{fig16} that many models added to the se module do not achieve better results. Due to the difference in datasets, when using the CK+ dataset, the feature map extracted by the network contains a relatively large number of effective information channels. The addition of the channel attention module not only increases the computational burden, but may also cause invalid feature selection. Therefore, whether to use channel attention depends on the specific dataset.

\section{Conclusion}\label{Conclusion}
In this paper, a new and efficient neural network model is proposed. It uses the spatial attention map we generate to learn the areas related to the target task. The introduction of spatial attention map provides more effective supervision information and pixel-level guidance for each layer of the network. The effectiveness of  spatial attention map can be proved by visualing the feature maps. In addition, we have carried out a large number of comparative experiments. The results show that our network still achieves better performance than traditional neural networks in the case of low depth. Using group convolution greatly reduces the parameters and FLOPs of the network model, and improves the running speed of the model.

\section*{Conflict of interest}
The authors declare that they have no conflict of interest.



\bibliography{mybibfile}

\end{document}